\documentclass{article}

\usepackage[final]{corl_2020} % Uncomment for the camera-ready ``final'' version.

\usepackage{amsmath}
\usepackage{amssymb}
\usepackage{booktabs}
\usepackage{multirow}
\usepackage{graphicx}
\usepackage{paralist}
\usepackage[font=scriptsize,labelfont=sl,labelsep=period]{caption}
\usepackage{wrapfig}
\usepackage{enumitem}
\usepackage[subtle]{savetrees}

\DeclareMathOperator*{\argmax}{\arg\!\max}
\newcommand{\ie}{i.e., }
\newcommand{\eg}{e.g., }

\usepackage[dvipsnames]{xcolor}
\usepackage{pifont}
\newcommand{\cmark}{\textcolor[HTML]{59a14f}{\ding{51}}}%
\newcommand{\xmark}{\textcolor[HTML]{e15759}{\ding{55}}}%

\title{\Large{Transporter Networks: Rearranging the Visual World\\for Robotic Manipulation}\vspace{-0.8em}} % \\\Large{for Robotic Manipulation}
\author{
  \textbf{Andy Zeng, Pete Florence, Jonathan Tompson, Stefan Welker, Jonathan Chien}\\
  \textbf{Maria Attarian, Travis Armstrong, Ivan Krasin, Dan Duong}\\
  \textbf{Ayzaan Wahid, Vikas Sindhwani, Johnny Lee}\\
  Robotics at Google
}

\begin{document}
\maketitle

%===============================================================================

\vspace{-2.5em}
\begin{abstract} %  allows for significantly more sample efficient learning of 
Robotic manipulation can be formulated as inducing a sequence of spatial displacements: where the space being moved can encompass an object, part of an object, or end effector.
% tompson_20201026 - nit: I understand that transporter networks is the plural of a single Transporter Network, which is the archiecture we're proposing, however it'd be nice to see consistency here with the title (which is "Transporter Networks"). It's slightly jarring to me that one is plural and the other isn't. Same for a few <-- Andy: tried "we propose Transporter Networks, using..." but wasn't sure if sounded great
In this work, we propose the Transporter Network, a simple model architecture that rearranges deep features to infer spatial displacements from visual input -- which can parameterize robot actions.
It makes no assumptions of objectness (e.g. canonical poses, models, or keypoints), it exploits spatial symmetries, and is orders of magnitude more sample efficient than our benchmarked alternatives in learning vision-based manipulation tasks: from stacking a pyramid of blocks, to assembling kits with unseen objects; from manipulating deformable ropes, to pushing piles of small objects with closed-loop feedback.
Our method can represent complex multi-modal policy distributions and generalizes to multi-step sequential tasks, as well as 6DoF pick-and-place.
Experiments on 10 simulated tasks show that it learns faster and generalizes better than a variety of end-to-end baselines, including policies that use ground-truth object poses.
We validate our methods with hardware in the real world. Experiment videos and code will be made available at \href{https://transporternets.github.io/}{\color{magenta}https://transporternets.github.io}
% Our model exhibits significant sample efficiency for vision-based manipulation without making any object-specific assumptions .
% The key idea is in the spatially-consistent overlaying of visual features, essentially casting visual manipulation as a spatially-consistent template-matching problem. 
% It is generally applicable to pick-and-place, and without modification is also capable of learning pushing tasks involving granular media and deformable objects.
% one-shot generalizes to simple tasks, and can accomplish tasks that require multi-step sequencing, 6-DOF pick-and-place, closed-loop visual feedback, and generalization to unseen objects.
%, and show that Transporter Networks offer benefits over the seemingly idealized scenario of knowing the ground-truth object poses for all objects. Pete: would love to work this in, but perhaps the real world mention is more important

\end{abstract}


%===============================================================================

\section{Introduction}

End-to-end models that map directly from pixels to actions hold the capacity to learn complex manipulation skills, but are known to require copious amounts of data \cite{levine2016end, kalashnikov2018qt}.
Integrating object-centric assumptions -- \eg object keypoints \cite{manuelli2019kpam,kulkarni2019unsupervised,nagabandi2020deep,liu2020keypose}, embeddings \cite{mees2017metric,jund2018optimization}, or dense descriptors \cite{florence2018dense,florence2019self,sundaresan2020learning} -- has been shown to improve sample efficiency \cite{florence2019self}. However, these representations often impose data collection burdens (i.e., configuring scenes with specific singulated objects) and still struggle to address challenging scenarios with unseen classes of objects, 
occluded objects, %non-rigid objects, 
highly deformable objects,
or piles of small objects \cite{suh2020surprising}.
This naturally leads us to ask: is there structure that we can incorporate into our end-to-end models to improve their learning efficiency, without imposing any of the limitations or burdens of explicit object representations?

In this work, we propose the Transporter Network, a simple end-to-end model architecture that preserves \textit{spatial structure} for vision-based manipulation, without object-centric assumptions:%. Its key aspects are:

\begin{itemize}[leftmargin=*]
  \item Manipulation involves rearranging things, which can be thought of as executing a sequence of spatial displacements: where the space being moved (\ie transported) can encompass an object(s), part of an object, or end effector. We formulate vision for manipulation as estimating these displacements. Transporter Networks directly optimize for this by learning to 1) attend to a local region, and 2) predict its target spatial displacement via deep feature template matching -- which then parameterizes robot actions for manipulation. This formulation enables high-level perceptual reasoning about which visual cues are important, and how they should be rearranged in a scene -- the distributions of which can be learned from demonstrations.
  \item Transporter Networks preserve the 3D spatial structure of the visual input. Prior end-to-end models \cite{levine2016end, kalashnikov2018qt} often use convolutional architectures with raw images, in which valuable spatial information can be lost to perspective distortions. Our method uses 3D reconstruction to project visual data onto a spatially-consistent representation as input, with which it is able to better exploit equivariance \cite{kondor2018generalization,cohen2016group} for inductive biases that are present within the geometric symmetries \cite{platt2019deictic} of the data for more efficient learning.
\end{itemize}

% that explicitly preserves the spatial structure of the visual input as a first-class citizen.
%Our key idea is that vision for manipulation can be formulated as estimating a sequence of spatial displacements: where the space being moved (\ie transported) can encompass an object, or the robot end effector, and can change over time.
%Our approach directly uses raw observations from 3D cameras as input, and makes no assumptions of objectness.
%By preserving spatial information, it is able to better exploit equivariance for inductive biases that are present within the symmetries of the data for more efficient learning.

In our experiments, Transporter Networks exhibit superior sample efficiency on a number of tabletop manipulation tasks that involve changing the state of the robot's environment in a purposeful manner:
from stacking a pyramid of blocks, to assembling kits with unseen objects; from manipulating deformable ropes, to pushing piles of small objects with closed-loop feedback.
Transporter Networks excel in modeling multi-modal spatial action distributions, and by construction generalize across rotations and translations of objects.
They do not require any prior knowledge of the objects to be manipulated -- they rely only on information contained within partial RGB-D data from demonstrations, and are capable of generalizing to new objects and configurations, and for some tasks, one-shot learning from a single demonstration.
%For some tasks, they have also shown to be capable of one-shot learning from a single demonstration.

%In summary, our main contribution is a new perspective on vision-based models for manipulation... the role of preserving spatial structure to improve end-to-end learning.
%We investigate its benefits in the context of vision-based manipulation, for which we present a simple model architecture that learns to attend to a local region and predict its spatial displacement, while retaining the spatial structure of the visual input.
%On 10 planar manipulation tasks, we demonstrate that Transporter Networks is a surprisingly effective approach to learning vision-based policies from demonstrations, and is orders of magnitude more sample efficient than other end-to-end alternatives.
%We also show that it has the capacity to learn 6DoF tasks...
%To facilitate further research in vision-based manipulation, our code, pretrained models, and simulation benchmarks with all tasks will be made open-source.

% Andy: took out itemized list to truncate a bit.
Our main contribution is a new perspective on the role of spatial structure and its capacity to improve end-to-end learning for vision-based manipulation.
We propose a simple model architecture that learns to attend to a local region and predict its spatial displacement, while retaining the spatial structure of the visual input.
On 10 unique tabletop manipulation tasks, Transporter Networks trained from scratch are capable of achieving greater than 90\% success on most tasks with objects in new configurations using 100 expert demonstrations, while other end-to-end alternatives struggle to generalize with the same amount of data.
%is a surprisingly effective approach to learning vision-based policies from a modest number of demonstrations, and is orders of magnitude more sample efficient than other end-to-end alternatives.
We also develop an extension to 6DoF tasks by combining 3DoF Transporter Networks with continuous regression to handle the remaining degrees of freedom.
To facilitate further research in vision-based manipulation, we plan release code and open-source Ravens, our new simulated benchmark with all tasks.
Ravens features a Gym-like API \cite{brockman2016openai} with a built-in stochastic oracle to evaluate the sample efficiency of imitation learning methods.

% \vikas{Great introduction! However, I'd recommend sharpening the statement of contributions with a few more impressive numbers from the experiments, e.g., outperforms such and such baselines by xx\% acrosss yy tasks.}

\begin{figure}[t]
\centering
  \includegraphics[width=\textwidth]{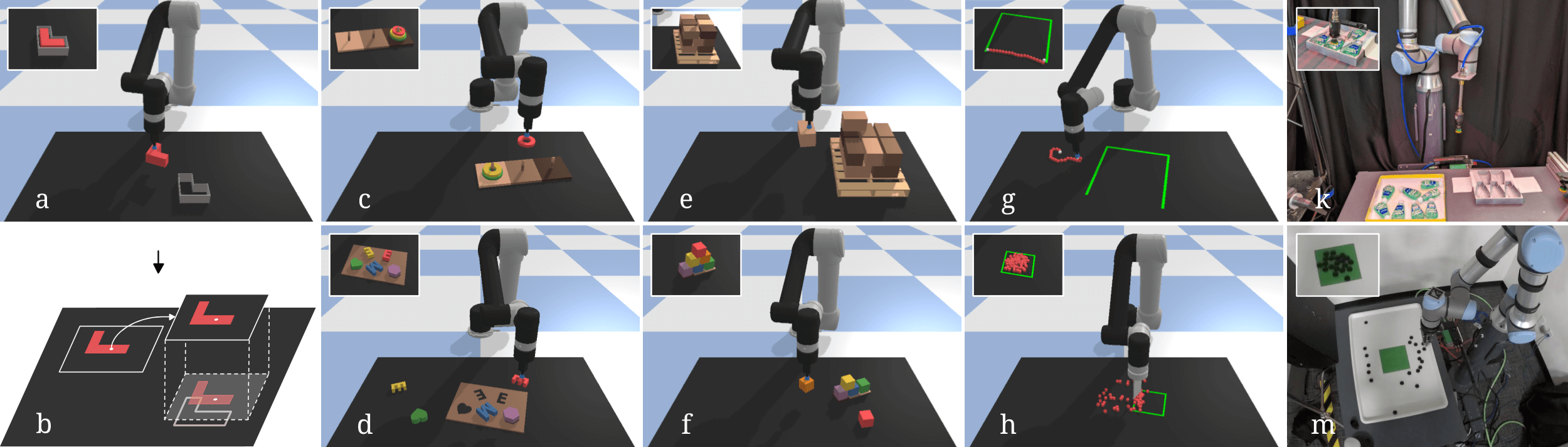}
  \caption{A Transporter Network is a simple model architecture that attends to a local region and predicts its spatial displacement (b) from visual input -- which can parameterize robot actions. It is sample efficient in learning complex vision-based manipulation tasks: inserting blocks into fixtures (a), sequential pick-and-place in Towers of Hanoi (c), assembling kits with unseen objects (d), palletizing boxes (e), stacking a pyramid of blocks (f), manipulating rope (g), and pushing piles of small objects with closed-loop feedback (h) -- and is practical to deploy on real production robots (k, m).}
  \label{fig:teaser}
  % https://docs.google.com/drawings/d/1tea6ZgI6kB2NdOatH7aVET-kyAmyxZVZpkExAeMKzxA/edit
  \vspace{-1em}
\end{figure}

\section{Related Work}

\textbf{Object-centric Representations.} Considerable research in vision for manipulation has been devoted to object detectors and pose estimators \cite{yoon2003real,zhu2014single,zeng2017multi,deng2019self,wang2019normalized}.
These methods often require object-specific training data, making them difficult to scale to applications with unseen objects.
Alternative representations including keypoints \cite{manuelli2019kpam,kulkarni2019unsupervised,nagabandi2020deep,liu2020keypose} and dense descriptors \cite{florence2018dense,florence2019self,sundaresan2020learning} have shown to be capable of category-level generalization and representing deformable objects, but still struggle to represent scenes with an unknown number of objects (\eg piles of small objects) or occluded objects. 
End-to-end models for vision-based manipulation learn faster when integrated with such object-centric representations \cite{florence2019self}, providing sample-efficiency benefits but retaining the data-collection and representational constraints of object-centricness.
%However, as a result they end up sharing the same generalization and data-collection constraints as their object-centric counterparts.
In this work, we show that it is possible to achieve sample efficient end-to-end learning without object-centric representations -- enabling the same model architecture to generalize to tasks with unseen objects, variable numbers of objects, deformable objects, and piles of small objects.

\textbf{Pick-and-Place} has had a long history in robotics, motivated by industrial needs.
Classic systems use pose estimation of known objects \cite{yoon2003real,zhu2014single}, as well as scripted planning and motion control \cite{frazzoli2005maneuver}.
While robust in structured settings (\eg manufacturing), these systems are difficult to deploy in unstructured settings (\eg logistics, agriculture, household) -- leading to renewed interest in leveraging learned models to achieve general pick-and-place policies \cite{gualtieri2018pick,zakka2019form2fit,hundtgood,berscheid2020self,wu2019learning,devin2020self,khansari2020action, song2020grasping} that can handle unseen objects.
We benchmark against the more recent Form2Fit \cite{zakka2019form2fit} on multiple tasks, and demonstrate that Transporter Networks are better able to handle tasks that require precise placing, multi-step sequencing, and closed-loop visual feedback.

\section{Method}

Consider the problem of learning pick-and-place actions $\mathbf{a}_t$ with a robot from visual observations $\mathbf{o}_t$:
$$f(\mathbf{o}_t)\rightarrow \mathbf{a}_t=(\mathcal{T}_\textrm{pick},\mathcal{T}_\textrm{place})\in\mathcal{A}$$
where $\mathcal{T}_\textrm{pick}$ is the pose of the end effector used to pick an object, and $\mathcal{T}_\textrm{place}$ to place the object.
Both poses can be defined in SE(2) or SE(3), depending on the task and degrees of freedom available to the end effector.
We can generalize beyond pick-and-place by considering manipulation with other motion primitives \cite{frazzoli2005maneuver} that are also parameterized by two end effector poses.
For example, we can parameterize pushing, where $\mathcal{T}_\textrm{pick}$ is instead the pose of the end effector at the start of the push, and $\mathcal{T}_\textrm{place}$ is where it moves to complete the push.
In this work, we consider tasks that can be completed by a sequence of two-pose motion primitives.
% While we focus on two-pose primitives, our approach could generally apply to longer sequences of pose changes $\{\mathcal{T}_\textrm{0},\mathcal{T}_\textrm{1},\mathcal{T}_\textrm{2},...\}$, which would be analogous to the continuous action parameterizations in prior work on end-to-end learning \cite{levine2016end}.
In the following section, we begin by describing our approach %Transporter Networks, 
in the context of planar pick-and-place.  
We then discuss extensions to 6DoF pick-and-place, sequential multi-step tasks, and motion-primitive-based manipulation of deformable objects and piles of objects.
%We then discuss its extensions to more degrees of freedom, followed by its application to other forms of manipulation. 

% some ideas for subfigures:
% (i) something that shows a picking heatmap distribution on the L object (you've already had this somewhere)
% (ii) something that illustrates the overlay-type operation (we had been prototyping these)
% (iii) something that shows a 3D view of the progression of the tower-of-hanoi task
% (iv) something that shows how in image space there are occlusions
% (v) something that shows a picking heatmap distribution on the tower-of-hanoi donuts too

\subsection{Learning to Transport}
%\subsection{Planar Pick-and-Place in SE(2)}
\label{sec:planar-pick-and-place}

\begin{wrapfigure}{r}{0.46\textwidth}
  \vspace{-1.7em}
  \begin{center}
    \includegraphics[width=0.46\textwidth]{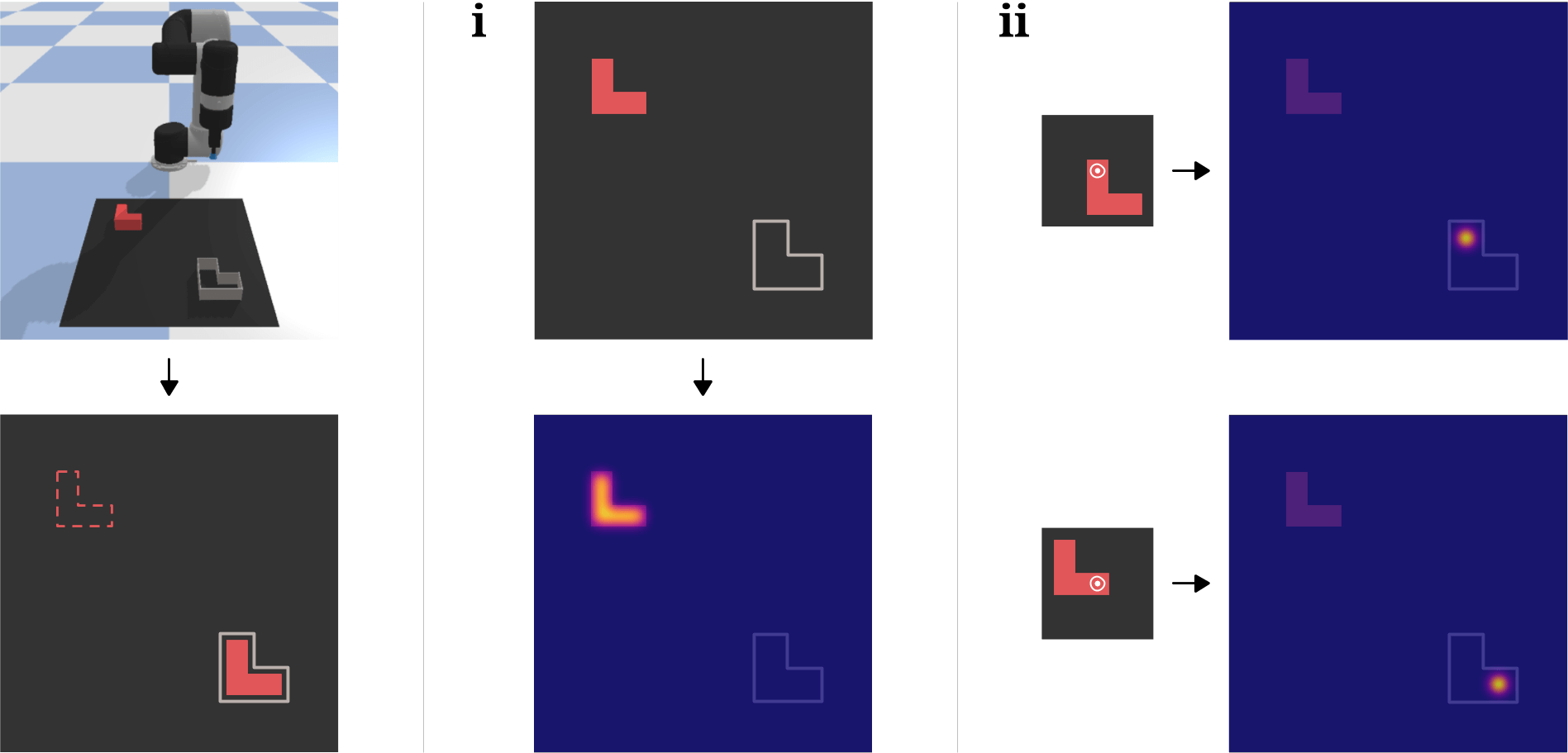}
  \end{center}
  \vspace{-0.6em}
  \caption{Simple planar pick-and-place task where (i) there is a distribution of successful pick poses, and (ii) for each successful pick pose, there is a corresponding distribution of successful place poses.}
  %\label{fig:block-insertion-easy}
  % https://docs.google.com/drawings/d/1cgvm7C6r5fgeBjrDKFGSPkuNkEYB8vV6tx_XCHZ_vEE/edit
  \vspace{-1em}
\end{wrapfigure}

Consider the planar pick-and-place problem on the right, where $\mathcal{T}_\textrm{pick},\mathcal{T}_\textrm{place}\in\mathbb{R}^2$ are 2D coordinates. In this task, the goal is to pick up the red block with an immobilizing grasp (\eg with a suction gripper), then place it into the fixture. This simple setting reveals two fundamental aspects of the pick-and-place problem: (i) there may be a distribution of successful pick poses (from which $\mathcal{T}_\textrm{pick}$ may be sampled \eg on the surface of the block), and (ii) for each successful pick pose, there is a corresponding distribution of successful place poses (from which $\mathcal{T}_\textrm{place}$ may be sampled).

The goal of Transporter Networks is to recover these distributions from visual observations alone -- without assumptions of objectness. While assigning canonical poses to the block and fixture might simplify the problem, there are many benefits to avoiding such object-centric representations -- including the capacity to handle unseen objects, deformable objects, and uncountable piles of small objects. Hence we assume no prior information about the objects (\eg no 3D models, poses, class categories, keypoints, etc.).

%A key idea in Transporter Networks can be appreciated in this simple example: what really matters to achieve a successful pick-and-place is the final alignment of the block into the fixture, and there is a multitude of $(\mathcal{T}_\textrm{pick},\mathcal{T}_\textrm{place})$ pairs that achieve this alignment.

%The key idea in Transporter Networks can be appreciated in this example by the combination of two simple observations: (i) there may be a distribution of pick poses ($\mathcal{T}_\textrm{pick}$) on the surface of the object, and (ii) corresponding to each successful pick pose, there is a correct place pose ($\mathcal{T}_\textrm{place}$) that successfully places the block into the fixture.
%The challenge that Transporter Networks addresses in particular is solving this problem from image observations only, without any canonical object pose data.  While the solution to this problem can be easily computed if the canonical poses of the block and fixture are available, there are many benefits to avoiding object-centricity and canonical object representations, for which the benefits (Section X) will be discussed.

% tompson_20201026: The notation here is a bit weird. p(x | y) is the conditional distribution over random variable 'x' conditioned on 'y'. In your notation, you're learning a function conditioned on the observation, but for some reason you use "|" with the pick prediction. I think it should be f_place(o_t, T_pick) (e.g. ',' not '|'). <-- Andy: yes great catch
Transporter Networks decompose the problem into (i) picking and (ii) pick-conditioned placing:
$$f_\textrm{pick}(\mathbf{o}_t)\rightarrow \mathcal{T}_\textrm{pick}\qquad\qquad\qquad\qquad f_\textrm{place}(\mathbf{o}_t,\mathcal{T}_\textrm{pick})\rightarrow \mathcal{T}_\textrm{place}$$
%We present a solution to both problems with Transporter Networks -- but in this work, our primary contribution is on the latter problem, pick-conditioned-placing. We begin by briefly describing picking first, followed by how Transporter Networks formulates placing.
%
We present a solution to both with Transporter Networks -- but in this work, our primary contribution is on (ii), pick-conditioned placing via transporting.  We first begin by briefly describing (i), picking.

\textbf{Learning Picking.} Our visual observation $\mathbf{o}_t$ is a projection of the scene (\eg reconstructed from RGB-D images), defined on a regular grid of pixels $\{(u,v)\}$ at timestep $t$ of a sequential rearrangement task.
% tompson_20201026: nit: is \eg required in the () above? I think you can just say (in the form of RGB-D images), since you don't consider any other alternatives. <-- Andy: changed this to \eg reconstructed from RGB-D images, since I think this could be heightmaps, or voxel grids, but not regular RGB-D images
Through camera-to-robot calibration, we can correspond each pixel in $\mathbf{o}_t$ to a picking action at that location: $\mathcal{T}_\textrm{pick}\sim(u,v)\in\mathbf{o}_t$.
The distribution of successful picks over pixels in $\mathbf{o}_t$ can be multi-modal in nature -- particularly in cases when there are multiple instances of the same object in the scene, or when there are symmetries in the shapes of objects. %of the object that allow multiple possible picking solutions.
As in prior works \cite{zeng2018robotic,morrison2018closing,zeng2018learning}, Transporter Networks use fully convolutional networks (FCNs, commonly used in vision for image segmentation tasks \cite{long2015fully}) to model an action-value function $\mathcal{Q}_\textrm{pick}((u,v)|\mathbf{o}_t)$ that correlates with picking success (architecture in Sec. \ref{sec:network-architecture}):
$$
\mathcal{T}_\textrm{pick} = \argmax_{(u, v)} \ \mathcal{Q}_\textrm{pick}((u,v)|\mathbf{o}_t)
$$
FCNs are translationally equivariant by nature, which synergizes with spatial action representations: if an object to be picked in the scene is translated, then the picking pose also translates.
%, which synergizes with spatial action representations like $\mathcal{T}_\textrm{pick}$ which can also represent a rigid transform applied to the input observation $\mathbf{o}_t$ to induce a spatial bias.
%Formally, equivariance here can be characterized as $\mathcal{Q}_\textrm{pick}(\mathcal{T}_\textrm{pick}\circ\mathbf{o}_t)=\mathcal{T}_\textrm{pick}\circ\mathcal{Q}_\textrm{pick}(\mathbf{o}_t)$
Formally, equivariance here can be characterized as $f_{\textrm{pick}}(g \circ \mathbf{o}_t) = g \circ f_{\textrm{pick}}(\mathbf{o}_t)$, where $g$ is a translation. Spatial equivariance has previously been shown to improve learning efficiency for vision-based picking \cite{khansari2020action,zeng2018robotic,morrison2018closing,zeng2018learning,zeng2019learning}. To this end, FCNs excel in modeling complex multi-modal picking distributions anchored on visual features.
%Picking distributions can also be non-isotropic, \eg the ring-shaped distribution for picking disks in Fig. \ref{fig:teaser}c.

% \vikas{nit: If you intend to be formal above, this equivariance definition intends to capture the idea that if the object to be picked in the scene is translated, then the pick point also translates. Correct? Then I would write this as $f_{pick}(g \circ o_t) = g \circ f_{pick}(o_t)$ where $g$ is a translation. I am fine with the above expression too if it is understood that  $\mathcal{T}_\textrm{pick}\circ\mathcal{Q}_\textrm{pick}$ denotes a new transformed action-value function. In other words, the transformation/group-action on a domain can be naturallly lifted to actions on functions over the domain.}  Andy: fixed! also added your example

% reference: On the Generalization of Equivariance and Convolution in Neural Networks to the Action of Compact Groups
% https://arxiv.org/pdf/1802.03690.pdf

\begin{figure}[t]
\centering
  \includegraphics[width=\textwidth]{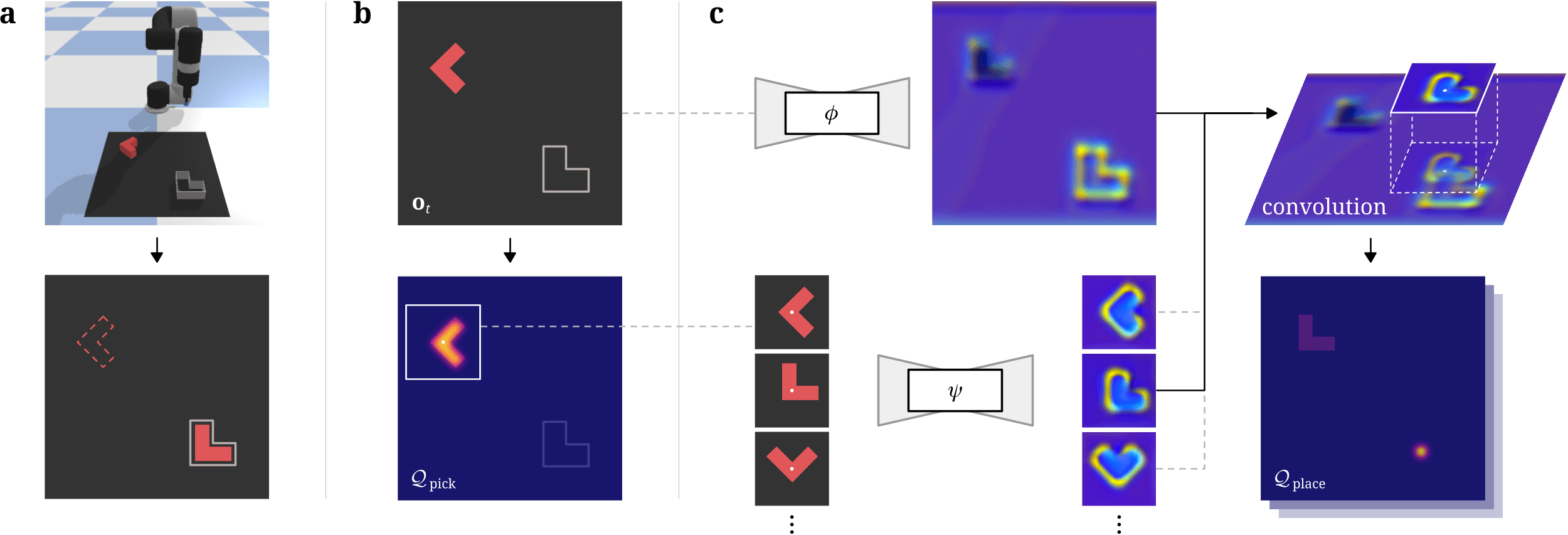}
  \caption{In this setting (a) where the task is to pick up the red block with an immobilizing grasp (\eg suction) and place it into the fixture, the goal of Transporter Networks is to recover the distribution of successful picks (b), and distribution of successful placements (c) conditioned on a sampled pick. For pick-conditioned placing (c), deep feature template matching occurs with a local crop around the sampled pick as the exemplar. Rotations of the crop around the pick are used to decode the best placing rotation. Our method preserves rotation and translation equivariance for efficient learning.}
  \label{fig:method}
  % https://docs.google.com/drawings/d/1TGGiSbjYEbUZ9kI3re9Kdly7R2-BWkphA9rGSNPY56k/edit
  \vspace{-1em}
\end{figure}

\textbf{Spatially Consistent Visual Representations.} The benefits of spatial equivariance are most pronounced when observations $\mathbf{o}_t$ are spatially consistent, meaning that the appearance of an object remains constant across different camera views, as well as across spatial transforms of rigid objects.
This is in contrast to perspective images, in which the appearance of an object could scale or distort depending on the camera view and lens properties.
In practice, spatial consistency also simplifies data augmentation during training, since applying rigid transforms to $\mathbf{o}_t$ can yield data that appears as if objects in the scene were physically reconfigured.
As with perspective images, however, spatially consistent representations still retain challenges from partial visibility and occlusion.
As in \cite{zeng2018learning}, we convert RGB-D images into a spatially consistent form by unprojecting to a 3D point cloud, then rendering into an \textit{orthographic} projection, in which each pixel $(u,v)$ represents a fixed window of 3D space (Fig.~\ref{fig:method}a).
%Fig. \ref{fig:method} illustrates an example of this with a top-down orthographic view of the block and fixture.
%However, spatially consistent representations generally make it easier for models to exploit geometric symmetries in the visible data. %, which appear more consistently than in perspective images.
Unlike in prior work, we use this spatially consistent representation for the \textit{transport} operation, which we describe next.

% \vikas{The concept of a metric space is very general, so I'd look for more precision above, unless its common 3D vision terminology. I have a naive question: Suppose I have two H x W x 3 arrays, one with RGB values and another with 3D coordinates. As data in $R^{2*(H x W x 3)}$, it lives in a metric space, but you dont mean that I presume.} Andy: changed to spatially consistent

\textbf{Learning Pick-Conditioned Placing via Transporting}.  Spatially consistent visual representations enable us to perform visuo-spatial \textit{transporting}, in which dense pixel-wise features from a partial crop $\mathbf{o}_t[\mathcal{T}_\textrm{pick}]$ centered on $\mathcal{T}_\textrm{pick}$ are rigidly transformed then overlaid on top of another partial crop $\mathbf{o}_t[\tau]$ centered on a candidate place pose $\tau$, where $\mathbf{o}_t$ is the observation before the pick.
Intuitively, if the crop encompasses an object, then this operation can be thought of as imagining the object at another location in the scene.
Fig. \ref{fig:method}c shows this in the context of overlaying the red block across the scene to find its placement in the fixture.

% \vikas{Clarification: $o_t$ is the observation acquired before the pick is executed, correct?}

In this work, % our query $\mathbf{o}_t[\mathcal{T}_\textrm{pick}]$ is a partial crop from $\mathbf{o}_t$ centered on $\mathcal{T}_\textrm{pick}$.
the goal of Transporter Networks is to transport our partial crop $\mathbf{o}_t[\mathcal{T}_\textrm{pick}]$ densely across a set of poses $\{\tau_i\}$ to search for its best placement, i.e. the $\mathbf{o}_t[\tau_i]$ with the highest feature correlation.
This operation is analogous to the search function in visual tracking literature \cite{li2019siamrpn++,bertinetto2016fully,tao2016siamese} where $\mathbf{o}_t[\mathcal{T}_\textrm{pick}]$ is the exemplar and $\mathbf{o}_t$ is the search area, or in retrieval systems literature where $\mathbf{o}_t[\mathcal{T}_\textrm{pick}]$ is the query and $\mathbf{o}_t$ is the key.
We formulate this as a template matching problem, using cross-correlation with dense feature embeddings $\psi(\cdot)$ and $\phi(\cdot)$ from two deep models:
% \pete{I think would be good to highlight that there is a pair of two networks.} Andy: added
\begin{equation}
\mathcal{Q}_\textrm{place}(\tau|\mathbf{o}_t,\mathcal{T}_\textrm{pick}) = \psi(\mathbf{o}_t[\mathcal{T}_\textrm{pick}])*\phi(\mathbf{o}_t)[\tau],\qquad
\mathcal{T}_\textrm{place} = \argmax_{\{\tau_i\}} \ \mathcal{Q}_\textrm{place}(\tau_i|\mathbf{o}_t,\mathcal{T}_\textrm{pick})
\label{eq:placing}
\end{equation}
where $\mathcal{Q}_\textrm{place}(\tau|\mathbf{o}_t,\mathcal{T}_\textrm{pick})$ is an action-value function that correlates with placing success, and $\tau$ covers the space of all possible placing poses -- which is equivalent to the space of 2D translations across $\mathbf{o}_t$ in our first simple red block and fixture example.  %We write $\Delta\tau$ with a $\Delta$ to emphasize this is best thought of as a relative displacement from the picking pose. %\footnote{In other notation, we may think of this as $\mathcal{T}^{\text{pick}}_{\text{place}}$}
%
% \vikas{Why are $\phi, \psi$ not the same since pick-and-place has reversible symmetry in a sense?}
% \pete{Due to the parameterized robot actions, the pick thing goes "on top" of the place thing, which to me breaks the symmetry and means we don't want $\phi, \psi$ to be the same.}
% \andy{We use two embeddings to avoid high correlation for the identity transform.}
%
%This simple matching function implies structure symmetry between the overlaid query and key features, \ie $\psi(\mathbf{o}_t[\mathcal{T}_\textrm{pick}])*\phi(\mathbf{o}_t[\Delta\tau])=\phi(\mathbf{o}_t[\Delta\tau])*\psi(\mathbf{o}_t[\mathcal{T}_\textrm{pick}])$.
Since $\mathbf{o}_t$ is spatially consistent, the structure of dense features $\psi(\mathbf{o}_t)$ and $\phi(\mathbf{o}_t)$ are also spatially consistent. % as $\Delta\tau$ covers Special Euclidean spaces SE(2) or SE(3).
A key property is that Eq.~\ref{eq:placing} is \textit{invariant} to $\mathcal{T}_\textrm{pick}$ as long as cross-correlation via transform $\mathcal{T}_\textrm{place}$ produces the desired imagined overlay.
%This is useful since picked objects can have features that align well with their target placement configurations, \eg when tightly packing boxes into a container, edges might align to edges, corners to corners, etc.
This enables learning pick-conditioned placing from few examples (Fig. \ref{fig:results}), without having to train over all possible picks \cite{wu2019learning}. Note that it is the deep features, not the RGB-D pixels, which are overlaid, and that the influence of the overlay extends beyond the crop, since for each dense pixel in feature space, the receptive field is large.

\textbf{Learning with Planar Rotations in SE(2).} Our first example uses only pick-and-place actions in the space of 2D translations -- however, $\mathcal{T}_\textrm{pick}$ and $\mathcal{T}_\textrm{place}$ may also represent rotations as in Fig. \ref{fig:method}.
In the planar SE(2) case, we discretize the space of SO(2) rotations into $k$ bins, then rotate the input visual observation $\mathbf{o}_t$ for each bin.
$\mathbf{o}_t$ is thereby defined on a grid of voxels $\{(u,v,w)_i\}$ where $w$ lies on the $k$-discretized axis of rotations, and $\mathcal{T}_\textrm{pick}$ and $\mathcal{T}_\textrm{place}$ are represented discretely on this grid in SE(2).
%Spatial equivariance can be preserved -- just as the weights of an FCN are shared across the space of translations, they can also be shared across rotations.
In practice, this is achieved by running the FCN $k$ times in parallel, once for each rotated $\mathbf{o}_t$, sharing model weights across all rotations.  
% pete: the next sentence is very interesting but it takes up 3 lines and we need to cut.
%Interestingly, adding rotation can subtly change the placing distributions -- for example, when placing disks for Towers of Hanoi in Fig. \ref{fig:teaser}c, the distribution of successful pick-conditioned placements is unimodal with translation-only placing, but spiral-shaped in SE(2) with rotations (See Appendix, Fig.~\ref{fig:towers-of-hanoi}d).
%pick-and-place for Towers of Hanoi in Fig. \ref{fig:towers-of-hanoi}d, we show how the solution space expands as $\mathcal{T}_\textrm{place}$ covers both rotations and translations.

% \vikas{Currrently Transporter Networks is trained on single step losses wrt ground truth placements. But one can also train it using a multi-step loss, no?}

% does not have to see everything

\label{sec:6dof-pick-and-place}
\begin{wrapfigure}{r}{0.5\textwidth}
  \vspace{-2em}
  \begin{center}
    \includegraphics[width=0.5\textwidth]{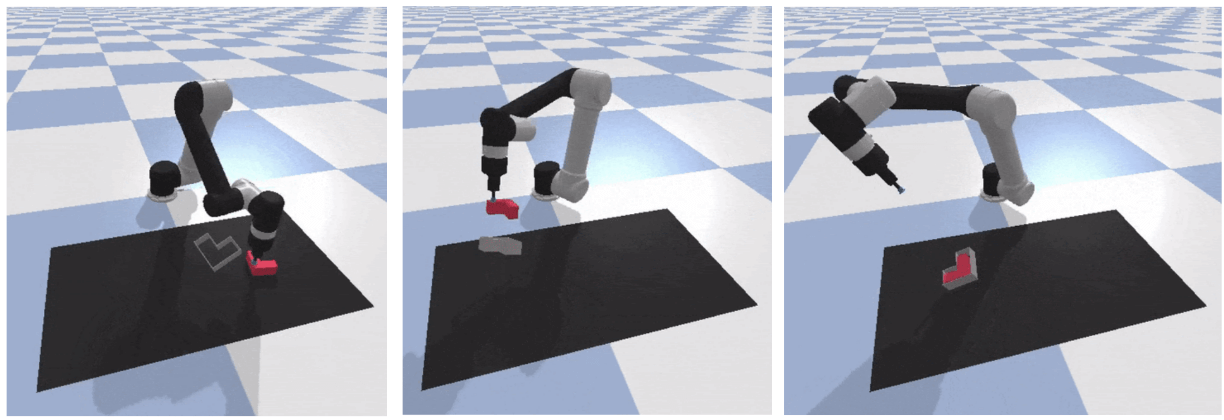}
  \end{center}
  \caption{In this 6-DoF variant of the task presented in Fig.~\ref{fig:method}, the fixture location varies as well in the out-of-plane rotations ($r_x$, $r_y$), and height ($z$).  Transporter Networks can address this task through multi-stage inference.}
  \label{fig:insertion-sixdof}
  % https://docs.google.com/drawings/d/1cgvm7C6r5fgeBjrDKFGSPkuNkEYB8vV6tx_XCHZ_vEE/edit
  \vspace{-1.0em}
\end{wrapfigure}
\textbf{Learning with Full Rigid Body Transformations in SE(3).} While the previous sections discussed learning a $\mathcal{Q}$ function over a discretized space, extending to the full SE(3) case to handle all six degrees of freedom of rigid objects becomes infeasible to sufficiently discretize the 6-dimensional space -- due to the curse of dimensionality.  However, interestingly we find the transport operation to be an enabling building block into higher dimensions, using a multi-stage approach to extend into tasks which require SE(3) placing.  
% tompson_20201026: nit: Slightly awkward sentence above. Suggestion: "However, interestingly we find the". <-- Andy: done
%
%Our SE(3) approach works as follows. 
First, the three SE(2) degrees-of-freedom are addressed as in the previous section, producing an estimated $\hat{\mathcal{T}}^{\text{SE(2)}}_{\textrm{place}}$.  Then, we extend the cross-correlation transport operation -- rather than using one channel over SE(2) to classify the correct discretized action, we use three channels over SE(2) to regress the remaining rotational ($r_x, r_y$) and translational ($z$-height) degrees of freedom.  We use a $\psi'$ and $\phi'$ which are identical to the $\psi$ and $\phi$ used for SE(2) classification, but with three separate cross-correlations ($*_3$) from subsets of channels, and also add learnable nonlinearity after the cross-correlation via a three-headed MLP network, $f(\cdot)$: $r_x, \ r_y, \ z = f \big( \psi'(\mathbf{o}_t[\mathcal{T}_{\textrm{pick}}])*_3\phi'(\mathbf{o}_t)[\hat{\mathcal{T}}^{\text{SE(2)}}_{\textrm{place}}] \big)$.
%
% \begin{equation}
% \begin{split}
% z &= f_z \big( \psi_z(\mathbf{o}_t[\hat{\mathcal{T}}^{\text{SE(2)}}_{\textrm{pick}}])*\phi_z(\mathbf{o}_t[\mathcal{T}_\textrm{place}^*]) \big)\\
% r_x &= f_{r_x} \big( \psi_{r_x}(\mathbf{o}_t[\hat{\mathcal{T}}^{\text{SE(2)}}_{\textrm{pick}}])*\phi_{r_x}(\mathbf{o}_t[\mathcal{T}_\textrm{place}^*]) \big)\\
% r_y &= f_{r_y} \big( \psi_{r_y}(\mathbf{o}_t[\hat{\mathcal{T}}^{\text{SE(2)}}_{\textrm{pick}}])*\phi_{r_y}(\mathbf{o}_t[\mathcal{T}_\textrm{place}^*]) \big)
% \end{split}
% \label{eq:regression}
% \end{equation}
%
% this is the actual one
% \begin{equation}
% \begin{split}
% r_x, \ r_y, \ z &= f \big( \psi'(\mathbf{o}_t[\mathcal{T}_{\textrm{pick}}])*_3\phi'(\mathbf{o}_t[\hat{\mathcal{T}}^{\text{SE(2)}}_{\textrm{place}}]) \big)\\
% \end{split}
% \label{eq:regression}
%\end{equation}
%
% But it would be nice to instead do this:
%
% \begin{equation}
% \begin{split}
% (z, r_x, r_y) &= f \big( \psi(\mathbf{o}_t[\mathcal{T}_\textrm{pick}^*]), \phi(\mathbf{o}_t[\mathcal{T}_\textrm{place}^*])\big)
% \end{split}
% \label{eq:regression}
% \end{equation}
%
This hybrid discrete/continuous approach affords a number of nice properties: the discrete SE(2) step is good at representing complex multi-modal distributions in those degrees of freedom, and once an approximate SE(2) alignment has been \textit{overlaid} via the transport operation, this acts as an attention mechanism to aid the model in precise continuous regression. To handle multi-modality in the continuous degrees of freedom, we can regress mixture densities.  In practice we have found that multiple approaches can work well once approximate SE(2) alignment has been applied, see the Appendix for more.  Multi-stage SE(3) classification, using Eq.~\ref{eq:placing} iteratively, is also possible.
%This hybrid discrete/continuous approach affords a number of nice properties: (i) the discrete SE(2) step is good at representing complex multi-modal distributions in those degrees of freedom, (ii) the remaining degrees of freedom out-of-image-plane are often unimodal once conditioned on the multi-modal in-plane distribution,  and (iii) once an approximate SE(2) alignment has been \textit{overlaid} via the transport operation, this acts as an attention mechanism to aid the model in precise continuous regression. In practice we have found that multiple approaches can work well once approximate SE(2) alignment has been applied, see the Appendix for more.  Multi-stage SE(3) classification, using Eq.~\ref{eq:placing} iteratively, is also possible.

%Our results show that at least under moderate out-of-plane rotations (+/- 45 degrees), the initial discrete step is able to classify accurately in SE(2) even though the additional degrees of freedom are not aligned.

%Where we have one MLP, and could maintain dimensionality before the MLP by having $*$ be per-channel.  In this case the input to $f$ would be $d-$dimensional, rather than $1-$dimensional.

%(Maybe this should go in Section 3.3) At training time, we use a Huber regression loss on each regression channel.

% don't think necessary and takes up a lot of space.
%\begin{equation}
%\mathcal{L}_{6DOF} = \mathcal{L}_{discrete} + \lambda_z \mathcal{L}_{z}(z, z^*) + \lambda_{r_x} \mathcal{L}_{r_x}(r_x, r_x^*) + \lambda_{r_y} \mathcal{L}_{r_y}(r_y, r_y^*) 
%\end{equation}

\textbf{Learning Sequential Pick-and-Place.} Consider the challenging example with 3-disk Towers of Hanoi in Fig. \ref{fig:teaser}c, where the task is to sequentially move 3 disks from the first tower to the third without placing a larger disk over a smaller disk.
Completing this task requires correctly sequencing 7 pick-and-place actions.
In this setup with a unique coloring of the base and disks, the task is Markovian.
Hence actions can be directly conditioned on visual contextual cues, \eg the next pick-and-place depends on the position (or absence due to occlusion) of observed objects in the scene.
Our method is stateless, but experiments show that it can learn such sequencing behaviors through visual feedback.
To enable this, it is important to increase the receptive field \cite{luo2016understanding} of the FCNs to encompass most of the visual observation $\mathbf{o}_t$. % A key aspect to enabling this behavior involves 
Extending Transporter Networks with memory to handle non-Markovian tasks would be interesting future work. % or leverage multi-step losses

\label{sec:beyond-pick-and-place}
\textbf{Manipulation Beyond Pick-and-Place: Deformables and Piles of Objects.}  Beyond pick-and-place of rigid objects, Transporter Networks can generalize to tasks that can be completed with two-pose primitives. For example, they can learn to sequentially rearrange a deformable rope such that it connects the two end points of an incomplete 3-sided square (Fig. \ref{fig:teaser}e), or sequentially push piles of small objects into a desired target set (Fig. \ref{fig:teaser}h) using a spatula-shaped end effector.
In both cases, our experiments (Sec. \ref{sec:experiments}) demonstrate that they can learn closed-loop behaviors using visual feedback to complete these tasks.
Both tasks are characterized by dynamics that cannot be modeled with rigid displacements -- however, our results suggest that rigid displacements can serve as a useful prior for non-rigid ones \cite{seita2020learning}.

% \vikas{Currently the paper is sharply focused on pick-and-place, and the generality discussed above is pitched as a "side-effect". I think this generality is very appealing and should be emphasized more} Andy: emphasized more in intro

\subsection{Network Architecture}
\label{sec:network-architecture}

\textbf{Observation Space}: For tabletop manipulation, our visual observation $\mathbf{o}_t$ is an orthographic top-down view of a $0.5{\times}1$m tabletop workspace, generated in simulation by fusing $480{\times}640$ RGB-D images captured with calibrated cameras using known intrinsics and extrinsics.
% tompson_20201026: For the above, I would clarify that this is only done for simulation. You later mention that multi-camera fusion is optional, but I'd make it clear that the real world experiments don't do it. <-- done.
The top-down image $\mathbf{o}_t$ has a pixel resolution of $160{\times}320$ -- each pixel represents a $3.125{\times}3.125$mm vertical column of 3D space in the workspace.
The image $\mathbf{o}_t$ contains both color (RGB) and scalar height-from-bottom (H) information, which enables the deep models to learn features that are rich in both visual texture and geometric shape.
% In total, $\mathbf{o}_t$ holds 4 channels of unique information (RGB-H) -- although empirically, we find that models that channel-wise clone the height images twice (for a total of 6 channels: RGB-HHH) are less likely to converge to a local optimum that uses only color.

\textbf{Picking}: Our picking model is a single feed-forward FCN that takes as input the visual observation $\mathbf{o}_t\in\mathbb{R}^{H{\times}W{\times}4}$ and outputs dense pixel-wise values that correlate with picking success: $\mathcal{V}_\textrm{pick}\in\mathbb{R}^{H{\times}W}=\textrm{softmax}(\mathcal{Q}_\textrm{pick}((u,v)|\mathbf{o}_t))$.
Our picking model is an hourglass encoder decoder architecture: a 43-layer residual network (ResNet) \cite{he2016deep} with 12 residual blocks and 8-stride (3 2-stride convolutions in the encoder and 3 bilinear-upsampling layers in the decoder), followed by image-wide softmax.
Each convolutional layer after the first is equipped with dilation \cite{yu2015multi}, and interleaved with ReLU activations \cite{nair2010rectified} before the last layer.
8-stride was chosen to balance between maximizing receptive field coverage per pixel prediction, while minimizing loss of resolution in the latent mid-level features of the network.
 
\textbf{Placing}: Our placing model is a two-stream feed-forward FCN that takes as input the visual observation $\mathbf{o}_t\in\mathbb{R}^{H{\times}W{\times}4}$ and outputs two dense feature maps: query features $\psi(\mathbf{o}_t)\in\mathbb{R}^{H{\times}W{\times}d}$ and key features $\phi(\mathbf{o}_t)\in\mathbb{R}^{H{\times}W{\times}d}$, where $d$ is the feature dimensionality.
Our placing model shares a similar hourglass encoder decoder architecture as the picking model: each stream is an 8-stride 43-layer ResNet, but without non-linear activations in the last layer.
A partial crop $\psi(\mathbf{o}_t[\mathcal{T}_\textrm{pick}])\in\mathbb{R}^{c{\times}c{\times}d}$ of query features with size $c$ centered around $\mathcal{T}_\textrm{pick}=\argmax{\mathcal{V}_\textrm{pick}}$ is
%indexed from the output of $\psi(\mathbf{o}_t)$,
transformed by $\tau$ (which covers the space of possible placement configurations), then cross-correlated (Eq. \ref{eq:placing}) with the key feature map $\phi(\mathbf{o}_t)$ to output dense pixel-wise values that correlate with placing success:
$\mathcal{V}_\textrm{place}\in\mathbb{R}^{H{\times}W{\times}k}=\textrm{softmax}(\mathcal{Q}_\textrm{place}(\tau|\mathbf{o}_t,\mathcal{T}_\textrm{pick}))$
where $\tau$ is discretized into $k=36$ angles (multiples of 10$^{\circ}$).
The translation and rotation $\tau$ which gives the highest correlation with the partial crop, via $\psi(\mathbf{o}_t[\mathcal{T}_\textrm{pick}])*\phi(\mathbf{o}_t)[\tau], $ %$\psi(\mathbf{o}_t[\mathcal{T}_\textrm{pick}])\in\mathbb{R}^{c{\times}c{\times}d}$ 
%with the highest correlation returns the best placement transform
gives $\mathcal{T}_\textrm{place}=\argmax{\mathcal{V}_\textrm{place}}$.
Note that this operation can be made fast if implemented with highly optimized matrix multiplication (where the crop is flattened) or as a convolution (where the crop is the convolving kernel).
Total inference time (with both picking and placing models) amounts to 200ms on an Nvidia GTX 2080 GPU.
\subsection{Training: Learning from Demonstrations}
\label{sec:training}
We assume access to a dataset $\mathcal{D}=\{\zeta_1,\zeta_2, ..., \zeta_n\}$ of $n$ expert demonstrations, where each trajectory $\zeta_i=\{(\mathbf{o}_0,\mathbf{a}_0), (\mathbf{o}_1,\mathbf{a}_1), ...\}$ is a sequence of one or more observation-action pairs $(\mathbf{o}_t,\mathbf{a}_t)$.
% tompson_20201026: Suggestion: "is a sequence of one or more successful observation-action pairs." (we don't use negative or noisy demonstrations).
In our experiments (Sec. \ref{sec:experiments}), we evaluate Transporter Networks in their ability to learn from $n$ = 1, 10, 100, or 1,000 demonstrations per task.
Our method does not take any task-specific information as input, and is trained on one task per model (future work may investigate multi-task).
%Our method does not make assumptions about which task it is trained for, nor does it take task-specific information as input.
%Its behavior is determined solely by the distribution of given demonstrations, which is disjoint per task in our experiments.
%
During training, we uniformly sample observation-action pairs from the dataset $\mathcal{D}$, from which each action $\mathbf{a}_t$ can be unpacked into two training labels: $\mathcal{T}_\textrm{pick}$ and $\mathcal{T}_\textrm{place}$, used to generate binary one-hot pixel maps $Y_\textrm{pick}\in\mathbb{R}^{H{\times}W}$ and $Y_\textrm{place}\in\mathbb{R}^{H{\times}W{\times}k}$ respectively. Our training loss is simply the cross-entropy between these one-hot pixel maps and the outputs of the picking and placing models: $\mathcal{L} = -\mathbb{E}_{Y_\textrm{pick}}[\textrm{log}\,\mathcal{V}_\textrm{pick}] -\mathbb{E}_{Y_\textrm{place}}[\textrm{log}\mathcal{V}_\textrm{place}]$.
% \begin{equation}
% \mathcal{L} = -\mathbb{E}_{Y_\textrm{pick}}[\textrm{log}\,\mathcal{V}_\textrm{pick}] -\mathbb{E}_{Y_\textrm{place}}[\textrm{log}\mathcal{V}_\textrm{place}]
% \end{equation}
While we only have labels for a single pixel per dense probability map, gradients are passed to all other pixels via image-wide softmax.
%Intuitively, this could be thought of as learning a soft ranking between pixels, where the logits belonging to pixels $\mathcal{T}_\textrm{pick}$ and $\mathcal{T}_\textrm{place}$ are pulled higher above others -- the scale of the gradients depends on the temperature of the softmax \cite{hinton2015distilling}.
We find this loss to be capable of learning multi-modal non-isotropic spatial action distributions.  For SE(3) models involving regression, we use a Huber loss on each regression channel.
% We find this image-wide softmax-based loss function to be capable of learning multi-modal non-isotropic spatial action distributions.  For SE(3) models involving regression, we use a Huber loss on each regression channel.
% We visualize several examples of these predicted modes in Sec. \ref{sec:modes}.
Validation performance generally converges after a few hours of training on a single commodity GPU.

% We later show that image-conditioned regression models (commonly used in prior work) may struggle to learn such distributions without more explicit techniques to control training data sampling.

% training details
% invariance to the coordinate frame in which data is presented

% runtimes
% 30ms preprocessing time
% In terms of runtime speeds, inference for the attention and transport models takes in total 200ms on an NVIDIA GTX 2080 GPU.

%===============================================================================

\section{Results}
\label{sec:experiments}

We execute experiments in both simulated and real settings to evaluate Transporter Networks across various tasks.
Our goals are threefold:
1) investigate whether preserving visuo-spatial structure within Transporter Networks improves their sample efficiency and generalization,
2) compare them to common baselines for end-to-end vision-based manipulation,
and 3) demonstrate that they are practical to deploy on real robots.

\subsection{Simulation Experiments}

\begin{wraptable}{r}{0.48\textwidth}
  %\begin{table} %[h]
  \setlength\tabcolsep{2.3pt}
  \vspace{-1em}
  \centering
  \scriptsize
  \begin{tabular}{@{}lcccc}
  \toprule
  & precise & multimodal & multi-step & unseen \\
  Task & placing & placing & sequencing & objects \\
  \midrule
  block-insertion (Fig. \ref{fig:method})   & \cmark & \xmark & \xmark & \xmark \\
  place-red-in-green                                 & \xmark & \cmark & \xmark & \xmark \\
  towers-of-hanoi (Fig. \ref{fig:towers-of-hanoi})   & \cmark & \cmark & \cmark & \xmark \\
  align-box-corner                                   & \cmark & \cmark & \xmark & \cmark \\
  stack-block-pyramid$^*$                            & \cmark & \cmark & \cmark & \xmark \\
  \midrule
  palletizing-boxes$^{*\mathsection}$                        & \cmark & \cmark & \cmark & \xmark \\
  assembling-kits$^{*\mathsection}$ \cite{zakka2019form2fit} & \cmark & \cmark & \xmark & \cmark \\
  packing-boxes$^{*\mathsection}$                            & \cmark & \cmark & \cmark & \cmark \\
  \midrule
  manipulating-rope$^{*\dagger}$                              & \cmark & \cmark & \cmark & \xmark \\
  sweeping-piles$^{*\dagger}$ \cite{suh2020surprising}        & \cmark & \cmark & \cmark & \xmark \\
  \bottomrule
  \end{tabular}
  \\$^*$tasks that have more than one correct sequence of states.
  \\$^\dagger$tasks that are beyond pick-and-place of rigid objects.
  \\$^\mathsection$tasks that are commonly found in industry.
  \caption{\scriptsize Each task in Ravens is characterized by a unique set of attributes.}
  \vspace{-1.8em}
  \label{table:task-attributes}
  %\end{table}
\end{wraptable}

We use behavior cloning simulation experiments to compare with baselines.  In this work, we do not use simulation for sim-to-real transfer -- rather only as a means to provide a consistent and controlled environment for fair comparisons and ablations. %Models evaluated in simulation are trained only on simulated data. Likewise models evaluated in the real-world are trained only on real data. % In simulation, we make no assumptions that cannot transfer to a real setup (\eg our method does not use ground truth state information).
Ravens, our simulated benchmark learning environment built with PyBullet \cite{coumans2016pybullet}, consists of a Universal Robot UR5e with a suction gripper overlooking a $0.5{\times}1$m tabletop workspace, with 3 simulated 640x480 RGB-D cameras pointing towards the workspace: one stationed in front (facing the robot), one on the left shoulder of the robot arm, and one on the right shoulder. Our method does not require 3 cameras -- having more cameras only improves visual coverage. For example, our real experiments only use 1 camera per robot. % of the workspace. %Each RGB-D image is rendered with salt-and-pepper noise via random Gaussian noise.
% tompson_20201026: Again, I'd point out that the real world experiments do not. <-- done

\textbf{Tasks and Metrics.} In simulation, we benchmark on 10 discrete-time tabletop manipulation tasks, some which require closed-loop visual feedback for multi-step sequencing.
We position-control the robot end effector (suction cup, or spatula for pushing) with two-pose motion primitives \cite{frazzoli2005maneuver}. %and IKFast \cite{diankov_thesis}.
For each task, all objects (including target zones) are randomly positioned and oriented in the workspace during training and testing.
%Hence, this experiment also tests generalization to new arrangements of objects.
%To succeed in the case where only a single demonstration is provided, the evaluated method needs to be invariant to unseen configurations of objects.
%To reflect the challenges of the real-world, observations contain only: raw RGB-D images and camera parameters (pose and intrinsics).
Each task comes with a scripted oracle that provides expert demonstrations by randomly sampling the distribution of successful $\mathcal{T}_\textrm{pick}$ and $\mathcal{T}_\textrm{place}$ -- samples of these are made available for the learner. %, but no other task information (objects, goal, etc.) is provided.
%No other information about the task is provided beforehand.
%For multi-step sequential tasks, information about the distribution of successful sequential solutions is only exposed through the distribution of demonstrations. % (\ie sequences of observations and actions).
Some tasks include multi-modality and permutations -- for example, in stacking, a successful 6-block pyramid is invariant to the permutation of blocks within each row, but not between rows. % (see individual task details in the Appendix). %The top row must always be a red block, middle must be orange or yellow, and the bottom must be green or blue or purple. %This information is only available from the distribution of demonstrations.
% More individual task details in the Appendix. % are in Sec. \ref{sec:task-details}.
Performance is evaluated with a metric from 0 (failure) to 100 (success). Partial credit is assigned during tasks that require sequencing multiple actions. We report results with the models that have achieved highest validation performance during training, averaged over 100 unseen test runs for each task. More details in the Appendix. % Sec. \ref{sec:evaluation-metrics}

\textbf{Baseline Methods.} %On behavior-cloning experiments in simulation, we compare with a number of relevant baseline models.
\textit{Form2Fit} \cite{zakka2019form2fit}, as mentioned in Related Work, is different from our method in that it does not perform visuo-spatial transporting%of partial crops from feature maps
, but instead uses a matching network. % to associate picking actions to independent placing actions.
%It consists of a picking network to predict candidate picking locations $\mathcal{T}_\textrm{pick}$, a placing network to predict candidate placing locations $\mathcal{T}_\textrm{place}$, and a matching network that computes dense visual descriptors \cite{florence2018dense,schmidt2016self} to associate picking actions to placing actions. 
% compute dense descriptors \cite{florence2018dense,schmidt2016self} to 
% \vikas{since it has similar components, it would help to add a sentence on differences between Form2Fit and Transporter.} Andy: done
\textit{ConvMLP} is a common model architecture in the end-to-end literature \cite{levine2016end} that uses a series of convolutional layers, followed by spatial-softmax and a multi-layer perceptron (MLP) -- here, we use this to regress the sequence of two SE(2) poses.  Form2Fit and ConvMLP use the same input images as Transporter Networks, but to get a sense of the difficulty of our tasks, we also supply two baselines which consume ground truth state (object poses and 3D bounding boxes) as input -- \ie assuming perfect object poses.
\textit{GT-State MLP} is an MLP which consumes these %all degrees of freedom of the environment as continuous values,
and regresses two SE(2) poses, while \textit{2-step GT-State MLP} first regresses the first SE(2) pose, then adds this to the observation vector and regresses the second SE(2) pose . To represent multi-modality, ConvMLP, GT-State MLP, and 2-step GT-State MLP regress mixture densities \cite{bishop1994mixture}. All methods, including ours, use identical data augmentation, with random SE(2) world-frame transforms.
%\andy{I think it'd be good to add back training from scratch, since it's usually the first thing people ask when I mention its sample efficiency}
%We do not use any pre-training from vision datasets (\eg ImageNet \cite{deng2009imagenet}) for our comparisons. All methods, including Transporter Networks, are trained from scratch. 

\textbf{Results: Sample Efficiency and Generalization on Benchmark Tasks}

Tab. \ref{table:sample-efficiency} shows sample efficiency of baselines trained from stochastic demonstrations for each task, and evaluated on unseen test settings, with random rotations and translations of objects (including target zones).
The benchmark is difficult -- most baselines, while capable of over-fitting to the demonstration training set, generalize poorly with only 1,000 demonstrations. In general, Transporter Networks achieve orders of magnitude more sample efficiency than the image-based alternatives, and also provides better sample efficiency than multi-layer perceptrons trained with ground truth state. %\pete{todo: a comment on this}
More analysis in the Appendix.
%\pete{TODO: comment more once table is full.}
%Form2Fit yields the highest performance over baselines on single-step tasks, but falls behind Transporter Networks on multi-step tasks that require closed-loop visual feedback.

\begin{table}[h]
  \setlength\tabcolsep{2.3pt}
  \centering
  \scriptsize
  \begin{tabular}{@{}lcccccccccccccccccccc@{}}
  \toprule
  & \multicolumn{4}{c}{block-insertion} & \multicolumn{4}{c}{place-red-in-green} & \multicolumn{4}{c}{towers-of-hanoi} & \multicolumn{4}{c}{align-box-corner} & \multicolumn{4}{c}{stack-block-pyramid} \\
  \cmidrule(lr){2-5} \cmidrule(lr){6-9} \cmidrule(lr){10-13} \cmidrule(lr){14-17} \cmidrule(lr){18-21}
  Method & 1 & 10 & 100 & 1000 & 1 & 10 & 100 & 1000 & 1 & 10 & 100 & 1000 & 1 & 10 & 100 & 1000 & 1 & 10 & 100 & 1000 \\
  \midrule
  Transporter Network                & \textbf{100} & \textbf{100} & \textbf{100} & \textbf{100} & \textbf{84.5} & \textbf{100} & \textbf{100} & \textbf{100} & \textbf{73.1} & \textbf{83.9} & \textbf{97.3} & \textbf{98.1} & 35.0 & \textbf{85.0} & \textbf{97.0} & \textbf{98.3} & 13.3 & \textbf{42.6} & \textbf{56.2} & \textbf{78.2} \\
  Form2Fit \cite{zakka2019form2fit} & 17.0 & 19.0 & 23.0 & 29.0 & 83.4 & \textbf{100} & \textbf{100} & \textbf{100} & 3.6 & 4.4 & 3.7 & 7.0 & 7.0 & 2.0 & 5.0 & 16.0 & \textbf{19.7} & 17.5 & 18.5 & 32.5 \\
  Conv. MLP                           & 0.0 & 5.0 & 6.0 & 8.0 & 0.0 & 3.0 & 25.5 & 31.3 & 0.0 & 1.0 & 1.9 & 2.1 & 0.0 & 2.0 & 1.0 & 1.0 & 0.0 & 1.8 & 1.7 & 1.7 \\
  GT-State MLP                        & 4.0 & 52.0 & 96.0 & 99.0 & 0.0 & 0.0 & 3.0 & 82.2 & 10.7 & 10.7 & 6.1 & 5.3 & 47.0 & 29.0 & 29.0 & 59.0 & 0.0 & 0.2 & 1.3 & 15.3 \\
  GT-State MLP 2-Step                        & 6.0 & 38.0 & 95.0 & \textbf{100} & 0.0 & 0.0 & 19.0 & 92.8 & 22.0 & 6.4 & 5.6 & 3.1 & \textbf{49.0} & 12.0 & 43.0 & 55.0 & 0.0 & 0.8 & 12.2 & 17.5 \\
  \midrule
  & \multicolumn{4}{c}{palletizing-boxes} & \multicolumn{4}{c}{assembling-kits} & \multicolumn{4}{c}{packing-boxes} & \multicolumn{4}{c}{manipulating-rope} & \multicolumn{4}{c}{sweeping-piles}\\
  \cmidrule(lr){2-5} \cmidrule(lr){6-9} \cmidrule(lr){10-13} \cmidrule(lr){14-17} \cmidrule(lr){18-21}
  & 1 & 10 & 100 & 1000 & 1 & 10 & 100 & 1000 & 1 & 10 & 100 & 1000 & 1 & 10 & 100 & 1000 & 1 & 10 & 100 & 1000 \\
  \midrule
  Transporter Network                & \textbf{63.2} & \textbf{77.4} & \textbf{91.7} & \textbf{97.9} & \textbf{28.4} & \textbf{78.6} & \textbf{90.4} & \textbf{94.6} & \textbf{56.8} & \textbf{58.3} & \textbf{72.1} & \textbf{81.3} & \textbf{21.9} & \textbf{73.2} & \textbf{85.4} & \textbf{92.1} & \textbf{52.4} & \textbf{74.4} & \textbf{71.5} & \textbf{96.1} \\
  Form2Fit \cite{zakka2019form2fit} & 21.6 & 42.0 & 52.1 & 65.3 & 3.4 & 7.6 & 24.2 & 37.6 & 29.9 & 52.5 & 62.3 & 66.8 & 11.9 & 38.8 & 36.7 & 47.7 & 13.2 & 15.6 & 26.7 & 38.4 \\
  Conv. MLP                           & 31.4 & 37.4 & 34.6 & 32.0 & 0.0 & 0.2 & 0.2 & 0.0 & 0.3 & 9.5 & 12.6 & 16.1 & 3.7 & 6.6 & 3.8 & 10.8 & 28.2 & 48.4 & 44.9 & 45.1 \\
  GT-State MLP                        & 0.6 & 6.4 & 30.2 & 30.1 & 0.0 & 0.0 & 1.2 & 11.8 & 7.1 & 1.4 & 33.6 & 56.0 & 5.5 & 11.5 & 43.6 & 47.4 & 7.2 & 20.6 & 63.2 & 74.4 \\
  GT-State MLP 2-Step                        & 0.6 & 9.6 & 32.8 & 37.5 & 0.0 & 0.0 & 1.6 & 4.4 & 4.0 & 3.5 & 43.4 & 57.1 & 6.0 & 8.2 & 41.5 & 58.7 & 9.7 & 21.4 & 66.2 & 73.9 \\
  \bottomrule
  \end{tabular}
  \vspace{0.5em}
  \caption{\scriptsize\textbf{Baseline comparisons.} Task success rate (mean \%) vs. \# of demonstration episodes (1, 10, 100, or 1000) used in training.}
  \vspace{-1.0em}
  \label{table:sample-efficiency}
\end{table}
% tompson_20201026: The above table is a lot to take in visually. Maybe consider bolding the best on each task?

\begin{wrapfigure}{r}{0.5\textwidth}
  \vspace{-0.6em}
  \begin{center}
    \includegraphics[width=0.5\textwidth]{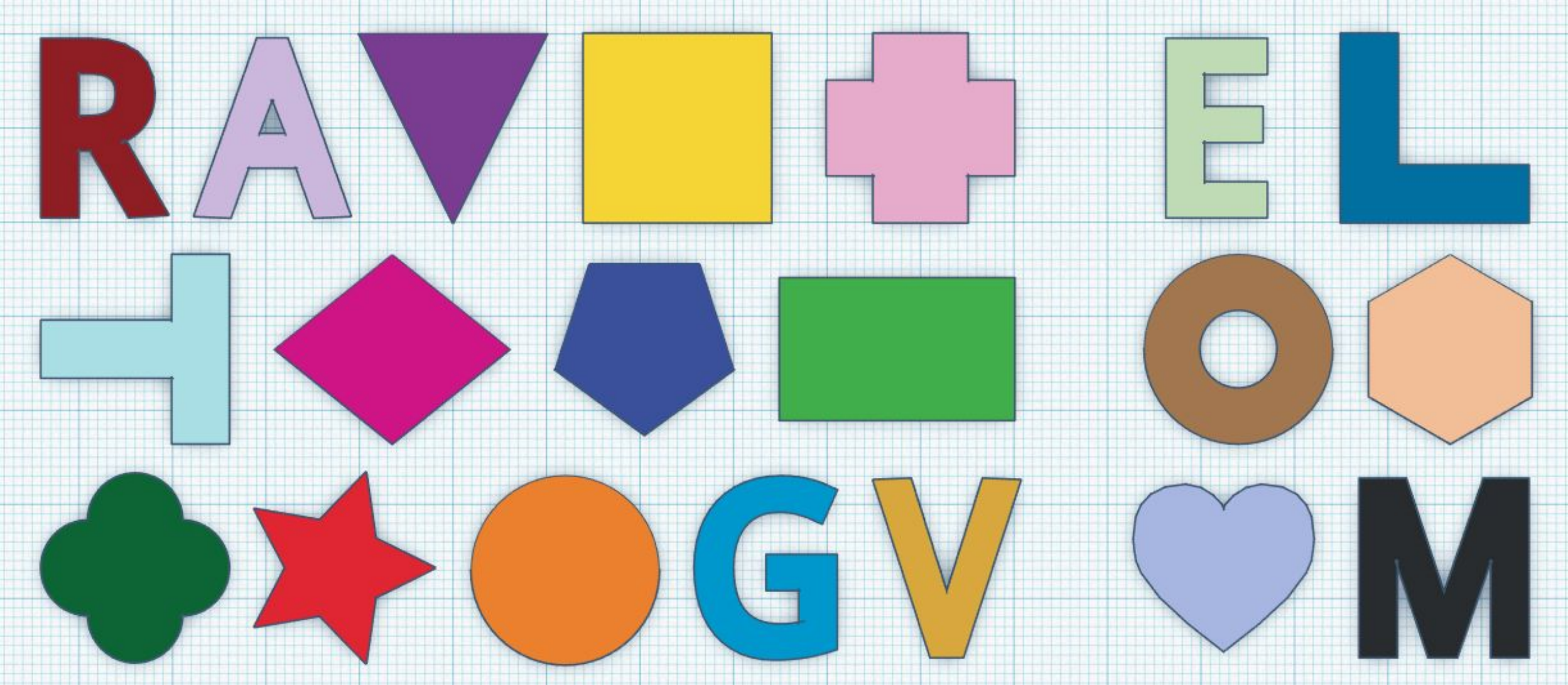}
  \end{center}
  \vspace{-0.5em}
  \caption{Kitting objects used in training (left 14) and testing (right 6).}
  \label{fig:kitting-objects}
  % https://docs.google.com/drawings/d/19rNXmjn_YsBrsWC_HwMTNGZz5H4Os211Aaey-N9lHTU/edit
  \vspace{-1em}
\end{wrapfigure}

\textit{Generalizing to Unseen Objects.} Three of our benchmark tasks involve generalizing to unseen objects (see Tab.~\ref{table:task-attributes}).
In our variant of the kit assembly task with unseen objects (shown in Fig. \ref{fig:kitting-objects}), the gap in performance between Form2Fit and Transporter Networks is rather large.
On a simpler set of unseen objects that have more primitive geometries (\eg circular disks, squares), which reflect the distribution of objects used in the original paper \cite{zakka2019form2fit}, Form2Fit achieves 96.3\% task success.
This suggests that Form2Fit descriptors have the capacity to express larger differences in geometry, but are less capable of matching higher resolution information than our deep template-matching based model, with the same amount of data.

\begin{figure}[t]
\centering
  \includegraphics[width=\textwidth]{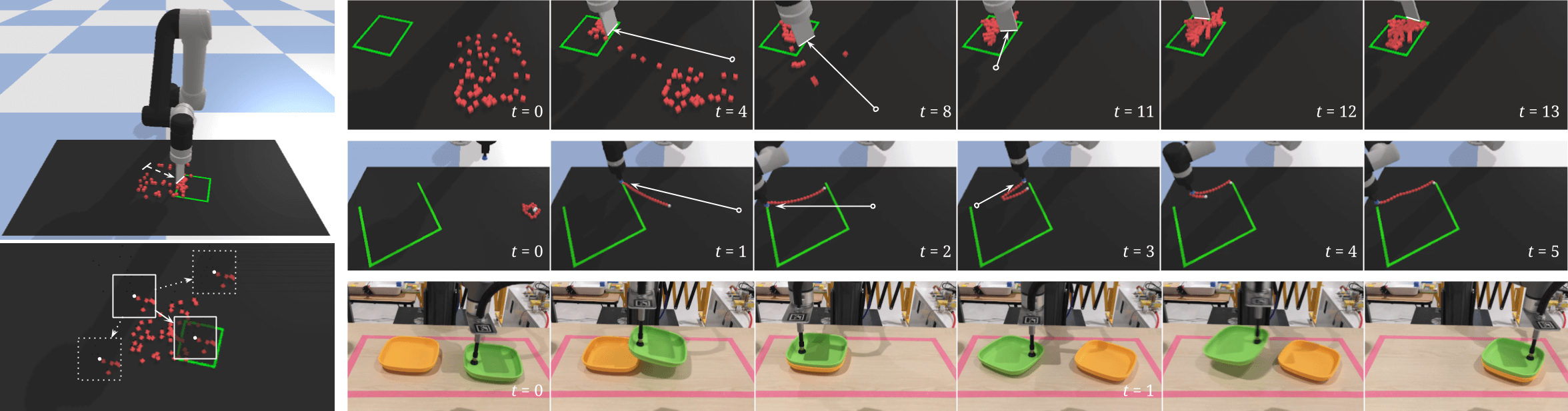}
  \caption{Transporter Networks can predict the desired spatial displacements of piles of small objects (left), which informs how to push them towards a target set. Our experiments show that it can learn visual feedback -- enabling it to recover from mistakes (\eg top row: pushing objects out of the target set at $t=11$), or sequentially rearrange a deformable rope (middle row). Since it learns pick-conditioned placing, it can also stack plates with varying initial predicted pick locations from only 10 demonstrations on a real robot (bottom row), without needing any object-specific representations.}
  \label{fig:results}
  % https://docs.google.com/drawings/d/1MvGvZj7oRpG_dxQ-P-jrwbqbymY0909dxcPHr2XYVxY/edit
  \vspace{-1em}
\end{figure}

\textit{Learning Sequential Manipulation with Closed-Loop Visual Feedback.} In this work, Transporter Networks are stateless models that react only to information presented as visual input during the current timestep.
%However, our experiments show that they have the capacity to mimic expert demonstrators with closed-loop behaviors.
However, our experiments show that the models have the capacity to learn visual feedback: they make use of contextual visual cues to determine which step of the task they are in, and use that to condition action-value predictions.
For example, when a stack of blocks falls over, they can re-build the stack of blocks as if they had just started the task.
If they accidentally push granular media out of the target zone during sweeping, they then push it back in (Fig. \ref{fig:results}).  We hypothesize that equivariance to rotations and translations enable them to learn these recovery behaviors even with little data on multi-step tasks (see Tab.~\ref{table:task-attributes}).
We show examples of these emerging behaviors in our supplementary video.

\begin{wraptable}{r}{0.47\textwidth}
  %\begin{table}[h]
  \vspace{-1em}
  \setlength\tabcolsep{2.3pt}
  \centering
  \scriptsize
  \begin{tabular}{@{}lcccccccc@{}}
  \toprule
  & \multicolumn{4}{c}{deterministic} & \multicolumn{4}{c}{stochastic} \\
  \cmidrule(lr){2-5} \cmidrule(lr){6-9}
  Method & 1 & 10 & 100 & 1000 & 1 & 10 & 100 & 1000 \\
  \midrule
  Transporter Network                & 100 & 100 & 100 & 100 & 100 & 100 & 100 & 100 \\
  Form2Fit \cite{zakka2019form2fit} & 100 & 100 & 100 & 100 & 100 & 30.0 & 45.0 & 45.0 \\
  Conv. MLP                        & 7.0 & 73.0 & 81.0 & 69.0 & 11.0 & 31.0 & 52.0 & 68.0 \\
  GT-State MLP                                 & 100 & 100 & 100 & 100 & 21.0 & 77.0 & 100 & 100 \\
  GT-State MLP 2-Step                          & 100 & 100 & 100 & 100 & 53.0 & 70.0 & 100 & 93.0 \\
  \bottomrule
  \end{tabular}
  \vspace{-0.5em}
  \caption{\scriptsize{Simplified 2DoF (no rotation) block-insertion is harder to learn with demonstrations from a stochastic oracle than a deterministic one.}}
  \vspace{-2.0em}
  \label{table:translation-only}
\end{wraptable}

\textit{Sample Complexity in Simplified Environments.}  Consider a simplified translation-only block-insertion task illustrated in Fig. \ref{fig:method}, where no rotations are needed, and the block is initialized to the same location, with only the fixture location varying between environments. % -- sampled variants of this environment have only 2DoF, as opposed to 6 in the original task, where both the block and fixture varies in SE(2).% $\mathcal{T}_\textrm{pick}$ and $\mathcal{T}_\textrm{place}$ .
We investigate two variants of experts in this setting: (a) one that provides deterministic demonstrations where $\mathcal{T}_\textrm{pick}$, $\mathcal{T}_\textrm{place}$ are fixed  relative to the block and fixture respectively, and (b) one that provides stochastic demonstrations where $\mathcal{T}_\textrm{pick}$, $\mathcal{T}_\textrm{place}$ are uniformly random sampled from the distribution of successful actions.
%Deterministic demonstrations are collected by an expert whose oracle , and $\mathcal{T}_\textrm{place}$ is fixed with respect to the pose of the fixture.
%In contrast, stochastic demonstrations are collected by an expert whose oracle %$\mathcal{T}_\textrm{pick}$ and $\mathcal{T}_\textrm{place}$ are randomly sampled from the distribution of successful actions.
Intuitively, the stochastic demonstrator reflects a more realistic setting in which the expert is inconsistent. %, and may provide a non-isotropic distribution of solutions as opposed to a single consistent solution at all times.
Tab. \ref{table:translation-only} displays sample complexity results on these tasks. %Transporter Networks against the baseline methods with deterministic and stochastic demonstrations.
Most baselines perform well with deterministic demonstrations, but begin to struggle with stochastic ones.
Transporter Networks, by construction, work well in both cases. We show additional extrapolation experiments in the Appendix.

\begin{wraptable}{r}{0.35\textwidth}
  %\begin{table}[h]
  \vspace{-1em}
  \setlength\tabcolsep{2.3pt}
  \centering
  \scriptsize
  \begin{tabular}{@{}lcccc@{}}
  \toprule
  & \multicolumn{4}{c}{6DoF block-insertion}  \\
  \cmidrule(lr){2-5}
            & 1 & 10 & 100 & 1000  \\
  \midrule
  Transporter Net. SE(3)               & 38.0 & 76.0 & 84.0 & 91.0 \\
  %GT-State MLP SE(2)          & 0.0 & 17.0 & 36.0 & 38.0 \\
  GT-State MLP 3-Step SE(3)        & 0.0 & 1.0 & 1.0 & 5.0 \\
  \bottomrule
  \end{tabular}
  \vspace{-0.5em}
  \caption{\scriptsize{6DoF block-insertion: our method generalizes with fewer demonstrations than GT-State MLPs.}} % task, success \% vs. $\#$ of demonstrations. 
  \vspace{-1.0em}
  \label{table:6dof}
\end{wraptable}

  % Task performance (mean \%) vs. \# of demonstrations (1, 10, 100, or 1000) used in training.
  
% \begin{wraptable}{r}{0.35\textwidth}
%   %\begin{table}[h]
%   \vspace{-1em}
%   \setlength\tabcolsep{2.3pt}
%   \centering
%   \scriptsize
%   \begin{tabular}{@{}lcccc@{}}
%   \toprule
%   & \multicolumn{4}{c}{6DoF block-insertion}  \\
%   \cmidrule(lr){2-5}
%   Transporter SE(3) (ours) & 1 & 10 & 100 & 1000  \\
%   \midrule
%   \% success               & -- & -- & -- & 91 \\
%   \hline
%   \textit{SE(2) Classification} \\
%   $x$/$y$ error, mean (mm)  & -- & -- & -- & -- \\
%   $r_z$ yaw error, mean ($^{\circ}$)  & 10* & 10* & 10* & 10* \\
%   \hline
%   \textit{Regression} \\
%   $z$ error, mean (mm)   & -- & -- & -- & -- \\
%   $r_x$ roll error, mean ($^{\circ}$)  & -- & -- & -- & 3* \\
%   $r_y$ pitch error, mean ($^{\circ}$)  & -- & -- & -- & 4* \\
%   \bottomrule
%   \end{tabular}
%   \vspace{-0.5em}
%   \caption{\scriptsize{6DoF block-insertion task showing per-DoF precision vs. $\#$ of demonstrations. }}
%   \vspace{-4.0em}
%   \label{table:6dof}
% \end{wraptable}

\textit{Learning 6DoF Pick-and-Place.}  We also test the SE(3) formulation discussed in Sec.~\ref{sec:6dof-pick-and-place} in the 6DoF environment shown in Fig.~\ref{fig:insertion-sixdof} with unseen target fixture configurations and stochastic demonstrations. On this more challenging task, our model demonstrates considerably better sample efficiency than the image-based baselines achieved on the strictly-easier 3DoF-only block-insertion task (Tab.~\ref{table:sample-efficiency} and \ref{table:6dof}). Additionally, while the hybrid discrete and continuous SE(3) model is not able to 1-shot generalize as the discrete-only SE(2) model could, we still observe substantially better performance than GT-State MLP models (Tab.~\ref{table:6dof}) which struggle in the larger observation and action spaces. For further analysis of this task and models, see the Appendix. % We find the MLPs struggle to precisely generalize with stochastic demonstrations
% SE2 more likley to recover in way that falls in distribution of training data

\subsection{Real-World Experiments}
\label{sec:real-world-experiments}

\begin{wraptable}{r}{0.33\textwidth}
  %\begin{table}[h]
  \vspace{-1em}
  \setlength\tabcolsep{2.3pt}
  \centering
  \scriptsize
  \begin{tabular}{@{}lcc@{}}
  \toprule
  Task & Test \% & \# Samples  \\
  \midrule
  Mouthwash Kit Assembly               & 98.9 & 8141 \\
  Sweeping Piles of Go Pieces        & 98.3 & 6759 \\
  \bottomrule
  \end{tabular}
  \vspace{-0.5em}
  \caption{\scriptsize{Transporter Networks test performance on real robots with human demonstration data.}} % task, success \% vs. $\#$ of demonstrations. 
  \vspace{-1.0em}
  \label{table:real}
\end{wraptable}

We validate Transporter Networks on both the kit assembly and the sweeping tasks (shown in Fig. \ref{fig:teaser}k and \ref{fig:teaser}m) using real UR5e robots.
For kit assembly, the robot uses a suction gripper and an industrial Photoneo PhoXi camera (for high resolution and accurate IR-depth).
For sweeping, the robot uses a brush end effector and a low-cost consumer-grade Azure Kinect (for RGB-depth).
Despite COVID-19 lockdowns preventing physical access, we were still able to perform real experiments using our Unity-based~\cite{unity2018unity} UI that enables people to remotely teleoperate our robots to calibrate, collect training data, and run tests.
% tompson_20201026: suggestion remove "(located domestically or internationally)". "remotely teleoperate" already captures this IMO. <-- Andy: agree
%Due to Covid-19 lockdowns, physical access to robots was limited.
%\pete{Maybe instead of: "due to covid, we made a unity UI."  could be instead, "despite covid-19 lockdowns, we were still able to collect real-world training data, since the training data can be teleoperated remotely.  we used a unity UI to..."} So we 
For these experiments, teleoperators are tasked with i) repeatedly assembling and disassembling a kit of 5 bottled mouthwashes, or ii) sweeping piles of small Go pieces into a target goal zone marked with green tape, and autonomous resets using a scripted scrambling motion.
% Kit assembly and disassembly are treated as separate tasks -- both use the same Transporter Network architecture with autonomous switching between the two tasks at inference time.
% a kit of 9 uniquely shaped wooden toys -- using a either a virtual reality headset, or mouse and keyboard to label picking and placing poses.
% Demonstrations are generally noisy, due to difficulty of the tasks (wood kits require 3-millimeter insertion precision).
% tompson_20201026: Is it only 3mm? It feels like only a 1mm on each side. Maybe you've measured it though. <-- Andy: yea Pete measured it to be 3mm with a caliper on his set
% To test the robustness of our method to diverse data sources, we also use both high quality depth camera observations (Photoneo PhoXi Model M) for one task, and those from low cost consumer-grade hardware (Azure Kinect RGB-D) for another.
%For this experiment, we train our models from noisy human remote teleoperation data as demonstrations.
%For demonstration data collection, we built a Unity-based UI to enable remote human teleoperators to calibrate and control the robot with a mouse and keyboard in order to label picking and placing poses.
%Teleoperators are tasked with repeatedly assembling and disassembling i) a kit of 5 mouthwashes, or ii) a kit of 9 uniquely shaped wooden toys. %The experimental setup of both tasks is shown in Figure \ref{fig:teaser}.
Table \ref{table:real} shows test performance of the models, and the amount of training data collected from 13 human operators in terms of transitions (\ie pick and place actions for kit assembly, pushes for sweeping).
% The models are trained with 11,633 pick-and-place actions total on all tasks from 13 human operators, and achieves 98.9\% success in assembling kits of bottled mouthwashes. % 
We also show qualitative results of our robots in action for both tasks in our supplementary video.
%Despite the noise in data, initial experiments suggest that Transporter Networks are capable of learning both tasks with on average 99\% picking success, 79.6\% placing success for assembling the kits (common failure: incorrect placement rotation), and 98.7\% placing success for disassembling the kits (common failure: stacking an object on another). <-- outdated numbers
%We also observe that placing precision is relatively sensitive to the accuracy of the calibrated camera extrinsics (due to compounding transforms). %camera-robot calibration
% Investigating improvements in real performance remains ongoing work.
% compounding
%99 picking
%79.6 kit (generous placing, 8 mouthwashes)
%98.7 dekit
% \andy{TODO: add observations of results}
% NOTE(tompson): The original text had 2,000 training samples, but we actually used a fair bit more than this:
% - F18DF2 / C6A model: 3492 training samples (not including validation)
% - reach12 model: 8141 training samples (not including validation)
% -- operators use a stopwatch-like interface to label the actions used for assembly, and those used for disassembly.
Further quantitative experiments were infeasible at time of submission due to COVID-19.
Additional experiment details in the Appendix.%; qualitative results of our robots in action for both tasks are shown in our supplementary video.
%\footnote{Extensive experimentation on real robots was limited due to Covid-19 lockdowns.}

% Highlight: 1) same architecture for dekitting / disassembly 2) difference between setups (diversity) 3) training from noisy (human) demonstrations. 

%===============================================================================

\section{Conclusion}
\label{sec:discussion}

In this work, we presented the Transporter Network, a simple model architecture that infers spatial displacements, which can parameterize robot actions from visual input. It makes no assumptions of objectness, exploits spatial symmetries, and is orders of magnitude more sample efficient in learning vision-based manipulation tasks than end-to-end alternatives: from stacking a pyramid of blocks, to assembling kits with unseen objects, to pushing piles of small objects with closed-loop feedback.
In terms of its current limitations:
% we have not demonstrated real-time high-rate feedback control,
it is sensitive to camera-robot calibration,
and it remains unclear how to integrate torque/force actions with spatial action spaces.
Overall, we are excited about this direction and plan to extend it to real-time high-rate control, and as well as tasks involving tool use.

%===============================================================================

% The maximum paper length is 8 pages excluding references and acknowledgements, and 10 pages including references and acknowledgements

\clearpage
% The acknowledgments are automatically included only in the final version of the paper.
\acknowledgments{%If a paper is accepted, the final camera-ready version will (and probably should) include acknowledgments. All acknowledgments go at the end of the paper, including thanks to reviewers who gave useful comments, to colleagues who contributed to the ideas, and to funding agencies and corporate sponsors that provided financial support.
Special thanks to Ken Goldberg, Razvan Surdulescu, Daniel Seita, Ayzaan Wahid, Vincent Vanhoucke, Anelia Angelova, for helpful feedback on writing, Sean Snyder, Jonathan Vela, Larry Bisares, Michael Villanueva, Brandon Hurd for operations and hardware support, Robert Baruch for software infrastructure, Jared Braun for UI contributions, Erwin Coumans for PyBullet advice, Laura Graesser for video narration.
}

%===============================================================================


\small
\bibliography{main}  % .bib
\normalsize

%===============================================================================

\clearpage
\section{Appendix}

The appendix consists of: descriptions of tasks and evaluation metrics in the Ravens simulation framework, additional experimental results, analysis, ablation studies, and real system details.

\subsection{Simulation Experimentation, Real World Experimentation, and COVID-19}

As introduced in the main paper, we execute experiments in simulation with Ravens as a means to provide a consistent environment for comparisons to baselines and ablations.
Models evaluated in simulation are trained only on simulated data. Likewise models evaluated in the real-world are trained only on real data.
In simulation, we avoid assumptions that cannot transfer to a real setup: observation data contains only 640x480 RGB-D images and camera parameters; actions are end effector poses (transposed into joint positions with inverse kinematics).
% tompson_20201026: "observation data contains"? Also, do you need "solvers" (e.g. "transposed into joint positions with inverse kinematics" seems sufficient). <-- done.
The only exception to this is our GT-State baseline, which learns MLPs that take ground truth state information (\eg object poses, bounding boxes, and color) as input -- shown to be less sample efficient than Transporter Networks for many tasks.
A limitation of Ravens is that the rendering software stack may not fully reflect the noise characteristics often found in real data: inaccuracies in camera calibration (\eg intrinsics and extrinsics), noise in commodity depth sensors. Hence, we also test our method on a real system (details in Sec. \ref{sec:real-experiments-details}) using teleoperated demonstration data, and present qualitative results in our supplementary video. Extensive quantitative experiments on real robots was difficult due to limited physical access during COVID-19.

\subsection{Ravens Simulation Framework}

In our simulation framework, Ravens, each task comes with a scripted oracle that provides expert demonstrations by randomly sampling the distribution of successful picking and placing actions -- samples of which are made available to the learner.
Ravens also has several attributes that make it useful for studying different areas that are beyond the scope of this work, including: 1) active learning, since algorithms can query an oracle during specific moments of uncertainty to improve learning efficiency, 2) reinforcement learning, since we provide reward functions that provide partial credit for all tasks (used only for evaluation in this work), and 3) active perception, since camera parameters (\eg extrinsics) are defined by an action (static in our experiments, but could be dynamic), which provides an opportunity to study algorithms that improve learning through active control of the camera position and orientation.

During both training and testing, all objects (including target zones) are randomly positioned and oriented in the workspace.
To succeed in the case where only a single demonstration is provided, the learner needs to be invariant to unseen configurations of objects.
Information about the multi-modality of a task or its sequential permutations is only made available from the distribution of demonstrations.
For example, moving disks in Towers of Hanoi may have only one correct sequence of states, but the distribution of how each disk can be picked or placed is learned from multiple demonstrations.
Or, when palletizing homogeneous boxes, the arrangement of boxes across each layer on the pallet must be transposed, and boxes should be stacked stably on top other boxes already on the stack to avoid toppling.
In our experiments, Transporter Networks are trained from scratch, without any pre-training from vision datasets. Exploring multi-task training for more efficient generalization to new tasks is interesting future work.

\begin{figure}[h]
\centering
  \includegraphics[width=\textwidth]{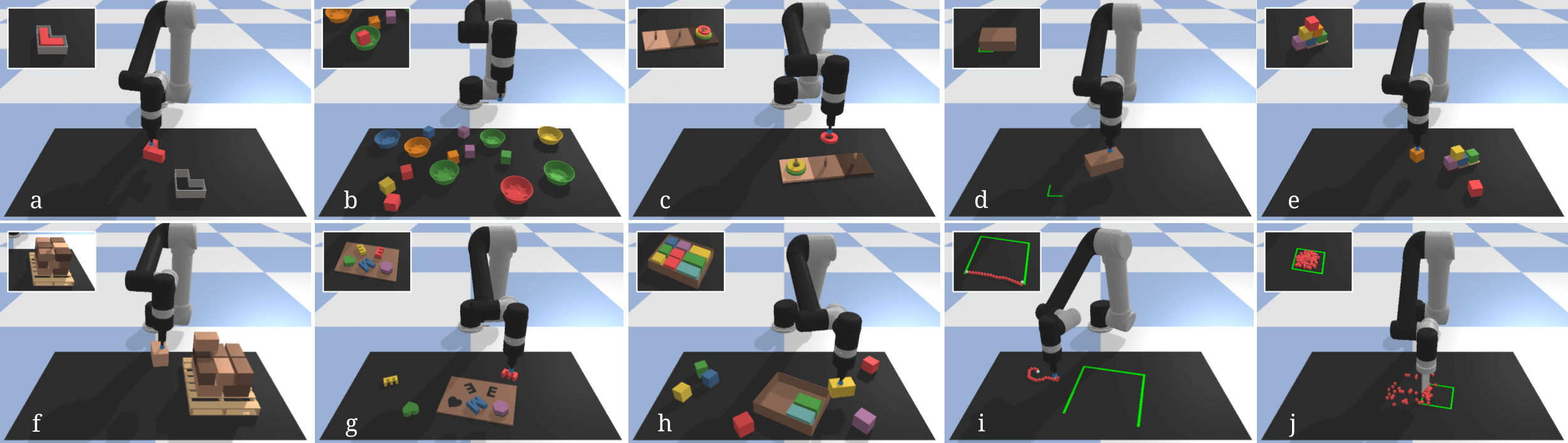}
  \caption{\textbf{Tasks:} (row-major order) block-insertion, place-red-in-green, towers-of-hanoi, align-box-corner, stack-block-pyramid, palletizing-boxes, assembling-kits, packing-boxes, manipulating-rope, sweeping-piles. Goal states (not provided to learners) are shown in top left corner of each image.}
  \label{fig:tasks}
  % https://docs.google.com/drawings/d/1q5zACf1JM53xG8hkvVIOLBaQ9DAsTaS2YRpEhqFxrgg/edit
\end{figure}

\subsection{Ravens-10 Benchmark Task Details}
\label{sec:task-details}

A set of 10 tasks (illustrated in Fig. \ref{fig:tasks}) comprise our Ravens-10 benchmark.  In each of these tasks, the agent acts with motion primitives parameterized by a sequence of two SE(2) end effector poses. Note that although the two motion-primitive-defining poses are in SE(2), manipulation behaviors that are in 3D can still be achieved (such as stacking blocks on top of each other) due to motion primitives that involve out-of-plane movement.

\textbf{block-insertion:} picking up an L-shaped red block and placing it into an L-shaped fixture with a corresponding hole. The block is 8cm in diameter while the hole is 9cm in diameter. %, which requires precise placing.
% tompson_20201026: I would remove "which requires precise placing." Some of your other tasks require much tighter than 1cm precision (e.g. 3mm precision mentioned earlier). Also, the definition of "precise" is relatively subjective. Alignment of lithography masks requires angstrom precision ;-) <-- Andy: agreed
The block and fixture configurations are randomly sampled in the workspace -- rejected if in collision with each other, or if partially out of bounds in the workspace.

\textbf{place-red-in-green:} picking up red blocks and placing them into green bowls amidst other visually distracting objects (\eg randomly positioned blocks and bowls of different colors). There can be multiple red blocks and/or multiple green bowls, with a random chance for more bowls than blocks. This task requires handling multiple candidate picking and placing locations. Since bowls are twice as large as the blocks, this task does not require precision. Red blocks and green bowls are first added to the workspace, then 10 objects (blocks or bowls of other colors) are randomly spawned in collision-free configurations.
  
\begin{figure}[h]
\centering
  \includegraphics[width=\textwidth]{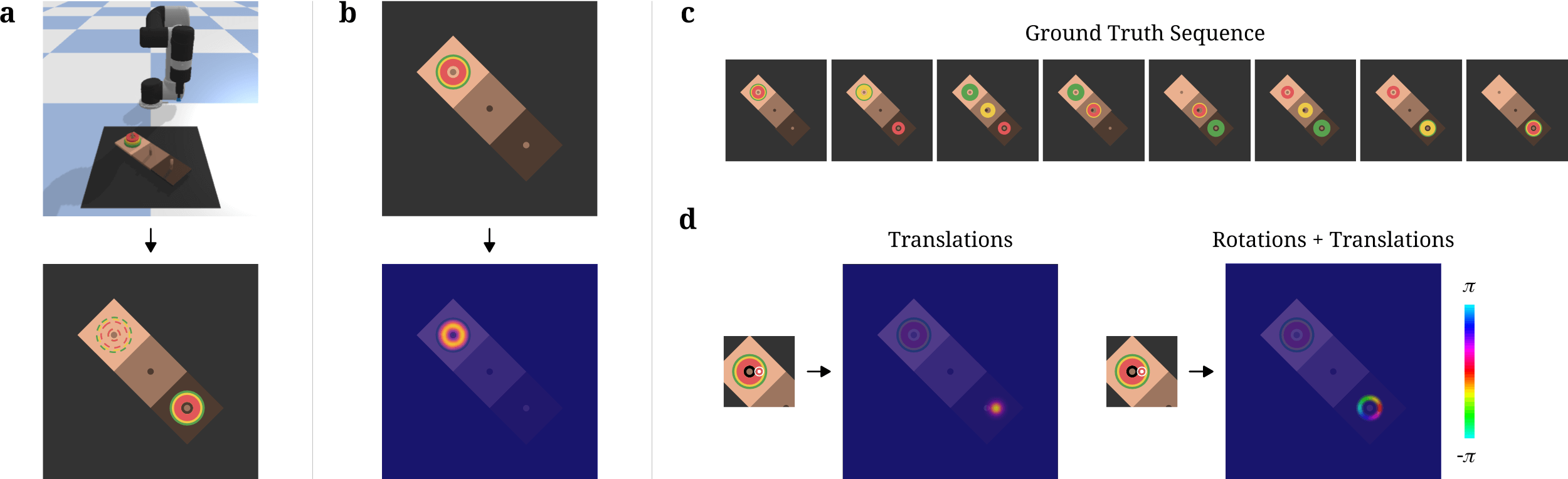}
  \caption{In Towers of Hanoi (a), the task is to sequentially pick-and-place 3 disks from the first peg to the third peg without placing a larger disk over a smaller disk (c). In each step, the distribution of successful picks is ring-shaped (b), and the distribution of successful pick-conditioned placements (d) is unimodal with translation-only placing, or spiral-shaped with SE(2) placing (projected onto a ring for visualization, where the ring traverses different rotations). The nature of this task evaluates the capacity of Transporter Networks to recover non-trivial spatial action distributions, and to sequence actions by leveraging contextual visual cues.}
  \label{fig:towers-of-hanoi}
  % \vspace{-0.5em}
  % https://docs.google.com/drawings/d/1YR7Q7fc42mkzmut-KUg_81E-ppNm3Kh-8OTl3B7jt9U/edit
\end{figure}

\textbf{towers-of-hanoi:} sequentially picking up disks and placing them into pegs such that all 3 disks initialized on one peg are moved to another, and that only smaller disks can be on top of larger ones. Completing this task requires sequencing 7 pick-and-places of the disks in a specific order. During evaluation, no information about which disk should be picked next is provided -- the approach needs to infer this from contextual visual cues \eg the positions or absence (due to occlusions) of other disks. This tests the capacity of the learner to infer the sequential semantics of Markovian vision-based manipulation tasks. Fig. \ref{fig:towers-of-hanoi} illustrates this task, its ground truth sequence, picking action distribution, and placing action distributions (unimodal for translation-only placing, or spiral shaped for the SE(2) placing). The base of the 3 pegs are randomly configured in the workspace, and disks are initialized on the peg on the lightest side of the base.

\textbf{align-box-corner:} picking up a randomly sized box and repositioning it on the tabletop to align one of its corners to a green L-shaped marker labeled on the tabletop. This task requires precision, and the ability to generalize to new box sizes. Aligning any of the box's corners to the green marker yields task success. The marker and box are randomly configured in the workspace such that the box does not overlap the marker. 

\textbf{stack-block-pyramid:} sequentially picking up 6 blocks (each with a different color) and stacking them into a pyramid of 3-2-1. The bottom 3 blocks can only be purple, green, or blue blocks. The middle 2 blocks can only be orange or yellow blocks. The top block can only be red. A successful pyramid is invariant to the permutation of blocks within each row, but not between rows. The blocks and base of the pyramid are randomly spawned in collision-free configurations.

\textbf{palletizing-boxes:} picking up fixed-sized boxes and stacking them on top of a pallet. To mimic real stability constraints, the arrangement of boxes should be transposed across each layer of the pallet. There are a fixed number of boxes -- each one is spawned into the workspace prior to each pick-and-place action.

\textbf{assembling-kits:} picking and placing different shaped objects onto corresponding locations on a board (visually marked with corresponding silhouettes of the objects). For each episode, 5 objects are randomly selected with replacement from a set of 20: 14 used during training (left), and 6 held out for testing (right). This task requires precise placing and generalizing to new objects by learning the concept of ''how things fit together" \cite{zakka2019form2fit}. The kit and 5 objects are randomly spawned in collision-free configurations.

\textbf{packing-boxes:} picking and placing randomly sized boxes tightly into a randomly sized container. This task is the hardest one in the benchmark, which requires not only relying on visual cues to tightly fit objects together, but also implicitly learning to order the objects in ways that maximize container coverage. During initialization, the container size is first randomly sampled, then randomly partitioned using a binary search tree (with minimum size of 5cm along each dimension). Each partition then parameterizes the size of each box, individually spawned into the workspace with a collision-free configuration.

\textbf{manipulating-rope:} manipulating a deformable rope such that it connects the two endpoints of an incomplete 3-sided square (colored in green). The cable is initialized as a randomly formed pile in the workspace, which needs to be untangled then aligned with the square. This task requires sequencing actions with closed-loop visual feedback to incrementally adjust the rope. The rope is implemented as a set of 20 articulated spherical beads, randomly dropped onto the workspace. The 3-sided square is randomly configured in the workspace.

\textbf{sweeping-piles:} pushing piles of small objects (randomly initialized) into a desired target goal zone on the tabletop marked with green boundaries. This sequential task (inspired by \cite{suh2020surprising}) requires sequencing pushes with closed-loop visual feedback and handling piles of small objects. In this task, the suction end effector is swapped with a spatula. The small objects and the target goal zone are randomly initialized in the workspace.

\subsection{Additional Tasks}

In addition to the Ravens-10 tasks, we used two other tasks in the paper for analysis.  The first was to provide a simplified test environment for testing the difference between stochastic and deterministic demonstrations (Sec. \ref{sec:experiments}), and interpolation/extrapolation experiments with a small enough amount of degrees of freedom such that they could be plotted (Sec. \ref{sec:interpolation-extrapolation}).  The second task was to test the SE(3) variant of Transporter Networks.

\textbf{2DoF simplified block-insertion:} This task is identical to the \textit{block-insertion} task, except: (i) the block is always in the same staring configuration, and (ii) the fixture varies only in translation.  The environment initialization has 2 degrees of freedom total.

\textbf{6DoF block insertion:} This task is identical to the \textit{block-insertion} task, except that the fixture varies in 6 degrees of freedom, requiring SE(3) placing. In addition to the table-plane degrees of freedom, the fixture also varies in height ($z$) as well as the two rotational degrees of freedom out of the plane, which may be represented by \textit{roll} and \textit{pitch}.  The block varies only in SE(2).  The environment initialization has 9 degrees of freedom total.

\subsection{Evaluation Metrics}
\label{sec:evaluation-metrics}

Performance on each task in simulation is evaluated in one of two ways:

\textit{Pose:} object translation and rotation to target pose is less than a threshold $\tau=1$cm and $\omega=15^\circ$ respectively. If completing a task requires sequencing multiple actions, then each successful action (where an object is moved to its correct pose) returns a partial reward $r=\frac{1}{\textrm{\# of actions}}$. Total rewards for completing a task always sums to 1. Object symmetries are accounted for. Tasks: block-insertion, towers-of-hanoi, place-red-in-green, align-box-corner, stack-block-pyramid, assembling-kits. 

\textit{Zone:} ratio of object(s) in the target zone. We discretize each object's 3D bounding box into 2cm$^3$ voxels. Total reward is the fraction of total voxels in the target zone $r=\frac{\textrm{\# of voxels in target zone}}{\textrm{total \# of voxels}}$. Tasks: palletizing-boxes, packing-boxes, manipulating-cables, sweeping-piles. Achieving total reward $r=1$ on palletizing-boxes and packing-boxes requires tightly fitting all objects into the target zone (\ie it is exponentially more difficult to achieve $r=0.9$ than it is to achieve the first $r=0.5$).

\subsection{Data Augmentation}
\label{sec:augmentation}

As is common with deep learning methods, we find that data augmentation can substantially benefit generalization performance for Transporter Nets.  To control the experimental comparisons with baselines in this paper, all methods used an identical data augmentation distribution.  Specifically, a random SE(2) transform, $T^{aug}_\mathcal{W}$, relative to the world frame $\mathcal{W}$ was sampled, and all observations (whether the images, or the object poses for the GT-State methods) were adjusted as if the data was observed from the augmentation frame $aug$ rather than the world frame.  This augmentation makes the assumption that the absolute position of objects on the table plane does not matter for the tasks -- this was indeed the case for all tasks shown.  Note that this distribution provides augmentation along the 3 degrees of freedom of SE(2), but we do not apply other augmentations such as imagining different relative configurations of objects -- this would not be a feasible assumption for certain tasks that involve multiple objects.

\subsection{Additional Baseline Details}

% Andy: probs don't need much on form2fit, paper has enough
%\textit{Form2Fit} \cite{zakka2019form2fit} is a recent method for pick-and-place that generalizes to new objects. It consists of a picking network to predict candidate picking locations $\mathcal{T}_\textrm{pick}$, a placing network to predict candidate placing locations $\mathcal{T}_\textrm{place}$, and a matching network that computes dense visual descriptors \cite{florence2018dense,schmidt2016self} to associate picking actions to placing actions. 
%Form2Fit is different from our method in that it does not perform visuo-spatial transporting of partial crops from feature maps.
%Instead, it uses its matching network to learn an embedding space where each pixel descriptor is optimized to represent the local visual region around it.
%As our experiments later show, training these descriptors can require substantially more data.
%While the scope of their experiments covers only one of our simulation tasks (assembling kits), their method could also be applied to other tasks -- which we compare to here.
%We use an author-provided implementation of Form2Fit with 43-layer ResNet backbones of similar parameter count for a fair comparison.

\textit{Form2Fit} \cite{zakka2019form2fit} is a recent method for pick-and-place (also similar to Devin et al. \cite{devin2020self}), using a matching network that computes dense visual descriptors \cite{florence2018dense,schmidt2016self} to associate picking actions to placing actions.  We use an author-provided implementation of Form2Fit with 43-layer ResNet backbones of similar parameter count for a fair comparison.

\textit{ConvMLP} is a model architecture first proposed in \cite{levine2016end} for learning control policies from pixel input -- commonly used in end-to-end literature. While typically this architecture is used for real-time continuous control and was originally used with a joint torque action space, we tried it here for our motion-primitive-based setting with a spatial action space. The input images are the same as that of Transporter Networks and Form2Fit: top-down reconstructed views from RGB-D data. The architecture consists of a series of three convolutional layers, followed by spatial soft(arg)max \cite{goroshin2015learning} and a multi-layer perceptron to regress a 6-dimensional vector of the two SE(2) end-effector poses. The architecture used was almost exactly as in \cite{levine2016end}, but with the following changes: (i) we used channel depths of 64, 64, 64, rather than the 64, 32, 16 used in the original paper, (ii) we used MLP layer widths of 128, 128, (iii) we used the MLP to regress mixture densities \cite{bishop1994mixture} rather than direct regression in order to handle multi-modality, and (iv) we added depth image input to the network, by replicating depth along 3 channels, processing it with one separate 7x7 convolutional layer, and then fusing with RGB after the first layer via concatenation.  The addition of depth image input into the model was to be more directly comparable with the other image-based options, which also used depth. As in \cite{levine2016end}, we use ImageNet-pretrained weights from GoogLeNet \cite{szegedy2015going} for the first RGB convolutional layer, which we find to aid in stable training and generalization. While this network is 10x shallower (3 layers) than the network used for Form2Fit and Transporter Nets (43 layers), we tried to train the same 43-layer backbone network used in Transporter Networks with a spatial soft(arg)max and MLP, but found that the shallower 3-layer network achieved better generalization at least for our tested hyperparameters.  Since many tasks involved multi-modal policies, it was necessary to train the network in a way that handles this multi-modality. Mixture Density Networks (MDN) \cite{bishop1994mixture} have previously been used for end-to-end manipulation control, for example in \cite{rahmatizadeh2018vision}.  As in \cite{bishop1994mixture}, we trained these with a negative log likelihood loss. We used a mixture of 26 Gaussians, with a training temperature of 2.5. To match translation and rotational units on a common scale, we equalized 1 $m$ translation with $0.1$ radians.  We also found improved generalization by taking the features $\mathbf{z}$ extracted by the spatial soft(arg)max, and concatenating $\mathbf{z} \leftarrow [\mathbf{z}, \ \sin(\mathbf{z}), \ \cos(\mathbf{z})]$ as input to the MLP -- this was inspired by \cite{mildenhall2020nerf,sitzmann2019siren}. We experimented with dropout in the MLP to add regularization, but found this to decrease generalization; the noise from the convolutional feature extraction was enough to regularize the MLP layers. The models were trained with the Adam optimizer for 20,000 steps, batch size 4, with a learning rate of $2e^{-4}$.

\textit{GT-State MLP} is a model architecture for learning control policies directly from ground truth state (i.e., object poses and 3D bounding boxes) as input, similar to the baseline used in \cite{florence2019self}. In the real world, this would be analogous to assuming perfect object pose estimation was provided by an external vision system.  The model shares an identical MLP to the ConvMLP model described above, except the input $\mathbf{z}$ to the model was all degrees of freedom of the environment (object poses, bounding boxes, and in the case of the place-red-in-green task we also added the RGB color of each object as input).  The bounding boxes are used for the 3 tasks (palletizing-boxes, align-box-corner, and packing-boxes) that involve boxes that vary over their length, width, and height, so the bounding boxes exactly parameterize these shapes. Some environments have a variable number of objects, and to handle this we padded the input vector with zeros to match the largest cardinality seen in the environment sampling.
We used dropout in the MLP layers with a dropout rate of $p=0.1$, which we found to improve generalization.  The models were trained with the Adam optimizer for 40,000 steps, batch size 128, with a learning rate of $2e^{-4}$.

\textit{GT-State MLP 2-step} is composed of two MLP networks, each of which are almost identical to GT-State MLP.  The first regresses only the first SE(2) pose, and the result of this is added to the observation vector of the second network, which regresses the second SE(2) pose. We used this model to experiment with whether the benefit of Transporter Networks was due to their sequential model cascade, where the second model is conditioned on the first (\ie pick-conditioned placing). We experimented with training the second MLP model on noisy regressions from the first model -- but perhaps because our models were sufficiently regularized with $p=0.1$ dropout, we found this to not provide measurable advantages over simpler teacher forcing, in which the second model was only trained on the conditioning from the training data.

\textit{GT-State MLP 3-step SE(3)} is almost identical to the 2-step model, except a third model is conditioned on the outputs of both the first and second model, and the additional out-of-plane degrees of freedom (height, roll, pitch) are regressed.  We experimented with also using just one GT-State MLP for the SE(3) action space task, but found similar performance.  Additionally, this model took in as input not only the SE(2) degrees of freedoms of objects, but their full SE(3) poses, represented with roll, pitch, yaw for the rotational degrees of freedom.  We experimented with other forms of rotational observation, such as observing three 3D points on the object, which does not have 2$\pi$ wrapping issues, but for the 6DoF task tested we found similar performance to the roll, pitch, yaw observation.

\subsection{Additional Analysis of Sample Efficiency and Baseline Comparisons}
\label{sec:additional-experimental}

Here we add additional analysis of the results in Tab. \ref{table:sample-efficiency} and Tab. \ref{table:translation-only}, which show the sample efficiency of baselines trained from stochastic demonstrations for each task, and evaluated on unseen test settings.

The results suggest that Transporter Networks accelerate end-to-end learning on the wide variety of tasks tested.  We believe this is due to the inductive bias of the model, which i) incorporates a high-level prior (that manipulation involves rearranging 3D space), and ii) uses the spatial structure of the visual input to exploit symmetries in the data to better learn useful correlations.

Among the two image-based baselines, Form2Fit performed better than ConvMLP on most tasks -- all except the sweeping-piles task, which may be especially hard for the descriptor-based Form2Fit model to work well on the pile of small items, since no item in the pile is significantly different than the others, but rather the policy should be a function of their collective distribution.  While for some tasks, like the place-red-in-green task, the Form2Fit model does comparably well to Transporter Networks, it struggles on tasks that require either more precision, or multi-step tasks. The Tab. \ref{table:translation-only} results for Form2Fit are also especially interesting -- although the model can 1-shot generalize with 100\% success on the simplified block insertion task, it starts to struggle fitting additional stochastic demonstrations.  The ConvMLP model for the most part sees a monotonic increase in performance with additional demonstrations, but often does not generalize as precisely as other models in this small-data regime. In all scenarios tested, Transporter Networks outperformed both image-based baselines.

For the state-based baselines, these experiments also show that Transporter Networks often require less data than MLP policies trained from ground truth state (\eg list of object poses, bounding boxes, and color). Of course, it is possible to have a policy that consumes the ground truth state of the scene and performs perfectly on all tasks -- indeed, the scripted expert policies do exactly this.  For learning generalizable policies from demonstration, however, our experiments suggest that directly reasoning over the spatial configuration of visual cues may instead provide grounds for discovering more generalizable patterns for manipulation.
The simplest example of this may be that it's easier to learn, from few demonstrations, a policy that is "pick a red pixel" and "place in a green pixel", rather than to regress these values in a continuous space, from a vector of continuous values.  Other additional cues that may be easier to learn directly from images include: aligning edges and corners when packing boxes and assembling kits, or matching gripper shapes to object surfaces for grasping.
%While ground truth state can deliver a concise description of change in the world, we conjecture that directly reasoning over the spatial configuration of low-level visual cues in the scene may.

In addition to the difference of using either images or ground-truth state, it is also important to keep in mind that the GT-State MLP model fundamentally differs in a couple other ways as well, including that its action space is continuous rather than discrete, and it is trained with a multi-modal regression loss rather than a classification loss.  It would be interesting to consider other model architectures that consume ground-truth state information and better model intra-object relationships and variable-dimensionality input, such as graph neural networks (GNNs), and also consider models that naturally include some translation equivariance, which MLPs do not.  Given the ubiquitity of MLPs in continuous regression tasks, however, and its proximity to the ConvMLP baseline, we consider our chosen MLP model to be a highly relevant baseline.

One additional note is that for the sweeping-piles task, we observe that a Transporter Network which is constrained to choose a sequence of two translation-only poses, where the rotation is determined as being orthogonal to the line between these two poses, works better than allowing the model to choose a rotation at each pose.  This may be due to the specifics of the sweeping motion-primitive used for this task, but it also may indicate that the rigid-transformation bias in the model has its limits.  The pile of objects does not get perfectly rigidly transformed, but rather has more complex dynamics.  We plan to investigate this further in future work.

\subsection{SE(3) Transporter Networks: Architecture Discussion and 6DoF Task Analysis}

\textbf{Network Architecture Considerations.}  We experimented with multiple versions of extending Transporter Networks into 6-degree-of-freedom, SE(3) action spaces.  The hybrid discrete/continuous model we have presented in the main paper provides a number of nice features, and demonstrates the flexibility of the transport operation.  It would also be interesting to try pure-discrete classification models, but fully sampling over the entire action space becomes burdensome with higher degrees of freedom. In the first SE(2) step, convolution can be used to efficiently imagine all overlays for the image-width and image-height dimensions, requiring sampling only over the in-plane rotations. However, the remaining 3 degrees of freedom in SE(3) are all orthogonal to the in-plane image dimensions.  While a discrete version could be done with rotating 3D voxel grids, this might present memory and sparsity challenges.  Instead, our presented choice of using continuous regression to directly regress the remaining degrees of freedom is an attractive one -- essentially building on the same architecture as the SE(2) version with little modification.

Here we provide additional details of our specific evaluated SE(3) approach with results shown in Tab. \ref{table:6dof}. As discussed in Section \ref{sec:6dof-pick-and-place}, all SE(2) degrees of freedom are resolved identically to the SE(2)-only models. An additional model then handles the remaining SE(3) degrees of freedom: roll, pitch, and $z$.  The additional model is almost identical to the SE(2) pick-conditioned placing model, but is trained to do three independent regressions rather than one classification.  In this way, it still benefits from the imagined overlay of the transport operation as an attention mechanism.  To adjust the architecture for this purpose, we first increased the output dimensionality to give $d=24$ channels, rather than the $d=3$ channels used for cross-correlation in the the classification models.  Additionally, rather than cross-correlating all 24 channels at once and reducing to an $\mathbb{R}^{H \times W \times k}$ tensor, for image height $H$, width $W$, and in-plane rotation discretizations $k$, we cross-correlate three distinct subsets of 8 channels to give an $\mathbb{R}^{H \times W \times k \times 3}$ tensor, which retains sufficient degrees of freedom to regress 3 independent values.  These 3 values in this tensor can be directly trained to do regression, but we find precision improvements if we add a small amount of non-linearity.  The small non-linearity mapping we use, $f(\cdot)$ in Sec. \ref{sec:6dof-pick-and-place}, is three independent MLP networks, which each take in one value and regress one value, and are each 3-layer, 32-width MLP networks with ReLu activations.  This is similar to 1x1 convolutions on the spatial feature maps, but with non-linearity.  We also find increased generalization if, rather than applying the regression loss only to the correct 1-hot label in $\mathbb{R}^{H \times W \times k}$, we also apply regression loss to a small window in that region.  We used a window of 7x7 pixels, and +/- 1 discretization in $k$.

We tried variations to the above described approach, many of which worked reasonably well.  One option, for example, is to use only one pick-conditioned placing network, which produces an $\mathbb{R}^{H \times W \times k \times 4}$ tensor, of which for the last tensor dimension, 1 channel is used for classification loss and the others are used for regression.  We found, however, both the classification loss and the regression loss to benefit from using separate networks--this is potentially due to differences in the training dynamics of the classification and regression losses.  We also tried various different methods of extracting the 3 regression channels from the spatial feature maps, including using spatial soft(arg)max from the overlaid crops followed by MLPs, and using concatenation rather than cross-correlation before the MLPs.  Each of these worked reasonably well.

\textbf{Additional 6DoF Task Performance Analysis.} Here we expand on analysis of the results presented in Tab. \ref{table:6dof}.  One important observation is that it was not clear, prior to these experiments, that the model would be able to successfully generalize due to its multi-step approach.  In particular, the SE(2) placing step occurs before control of the remaining degrees of freedom.  While this might be more challenging for different objects such as long and thin ones, we find that for our tested environments, we are able to get above-90\% generalization on this challenging task given 1,000 training examples.  Introspection into the accuracy of the different degrees of freedom of this model provides interesting comparisons as well.  Specifically, for $n=1,000$ training examples, the mean absolute errors (MAE) for roll/pitch regressions were each in the range of $3$ to $4$ degrees, whereas the MAE for in-image-plane rotation (yaw) were in the range of $10$ degrees, and most often contributed to unsuccessful deployments in test environments.  This suggests that future work building on this model could investigate an iterative approach in which the initial degrees of freedom would once again be optimized, potentially iteratively $N$ times, and potentially with increasingly fine resolution discretization.  Additionally, it should be noted that in our tested model the discrete step occurs before the regression step, and so in a sense has a more challenging task, since there may not be good alignment of the remaining degrees of freedom.  %Also, the discrete components of the model are more data-efficient than the continuous components -- as shown in Tab. \ref{table:sample-efficiency} the SE(2) model can fully solve this task from a single demonstration.
Additionally, we only used SE(2) augmentation as discussed in Sec. \ref{sec:augmentation}, since the presence of gravity does not make all roll/pitch-augmented worlds equal, and so the remaining SE(3) degrees of freedom do not benefit from the same level of augmentation.  We do find however that even in the $n=1$ demonstrations case, with 38 \% success, the SE(3) model behaves as if it uses one azimuth angle, consistent with a single roll/pitch being rotated around the $z$ axis, but this is actually a manifold of different roll/pitch values.  This shows the benefit of SE(2) augmentation even for the remaining SE(3) degrees of freedom.

Also as shown in Tab. \ref{table:6dof}, our tested GT-State MLP model significantly struggled with the 6DoF SE(3) task, much more so than it did with the 3DoF SE(2) task. Investigation into the cause of the challenge shows that this is primarily due to the additional degrees of freedom in the observation and action spaces.  In particular we can see this in the tested 3-step model (described in Sec. \ref{sec:additional-experimental}), which often struggles to precisely regress the $x$ and $y$ translational degrees of freedom even with $n=1,000$ demonstrations, which was not a problem in the simpler task.  Additionally, if we take this task's data, but reduce the observation and action spaces to only be SE(2), then with $n=1,000$ demonstrations the model is once again able to precisely regress the $x$ and $y$ translations.  We hypothesize that the mix of the stochastic demonstrations and the full 6-DoF placing is challenging for the model to learn a generalizable mapping with this amount of data.

\subsection{Interpolation and Extrapolation}
\label{sec:interpolation-extrapolation}

\begin{figure}[t]
\centering
  \includegraphics[width=\textwidth]{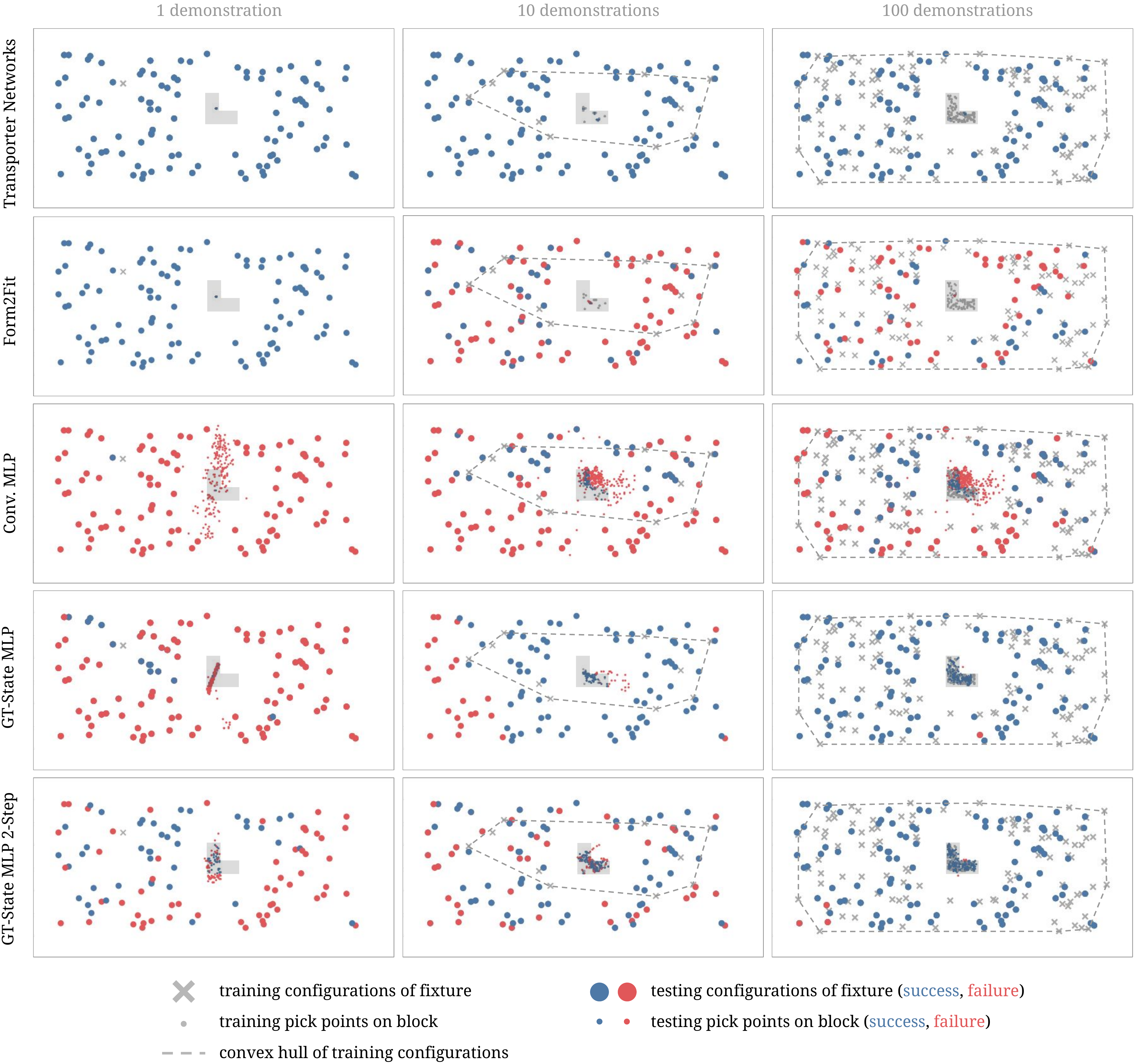}
  \caption{Depictions of the generalization ability of different models on the simplified translation-only block-insertion task. Each episode is visualized as a mark on the workspace (representing the pose of the fixture) -- color-coded by task success (blue for success, red for failure).  Environment configurations in the training set are shown in gray, and the convex hull of the training samples are drawn with a dotted line. Testing episodes inside the convex hell represent interpolation; those outside represent extrapolation.  Plot inspired by \cite{florence2019self}.}
  \label{fig:interpolation-extrapolation}
  % https://docs.google.com/drawings/d/1OEUQQd-FA2o4Gz5Tc4Dggiw1uoJi3eUp9lUCYNqQF1k/edit
\end{figure}

To gain a better qualitative understanding of the generalization abilities of Transporter Networks and the tested baseline models, we provide
Fig. \ref{fig:interpolation-extrapolation}. Similar to the analysis in \cite{florence2019self}, this shows the spatial distribution of training configurations and successful/unsuccessful testing configurations--in this case, for translation-only block insertion with stochastic demonstrations.  For each method we visualize its ability to interpolate and extrapolate from its training samples to handle new unseen poses of the target fixture.

As can be seen from these generalization plots, on this task Transporter Networks is capable of extrapolating to new testing episodes with just one demonstration. The dense deep feature transport operation enables the optimal placement of the block in the fixture to look similar under different picking and placing actions (exploiting translational equivariance of the task). This generalizes to new object configurations better than Conv MLP, which seems to only successfully interpolate around the majority of training episodes (often collapsing towards one mode of the training data).

Form2Fit is also capable of extrapolating from a single example, since both its picking and placing networks (also modeled by FCNs) are translationally equivariant -- such that the block insertion task amounts to detecting the block and fixture under different translations. %Its matching network descriptors do not need to be optimized for, as long as the picking network predicts the same pick on the block, and the placing network predicts the same place in the fixture. 
Once the number of demonstrations increase, however, the matching network is forced to learn pixel-wise descriptors that index multiple possible picking actions to multiple possible placing actions. In the setting with stochastic demonstrations, the positive-pair samples available for training the matching network are sparse and limited only to the pick-and-place action pairs from demonstrations -- making it difficult to geometrically regularize the descriptors uniformly over the pixels on the surface of the block and the fixture. Minor mis-matches during testing can also lead to a slight misalignment of the block, which leads to poor placement and task failure. This suggests that Form2Fit's matching network requires substantial training data to successful converge to a meaningful solution. The distribution of successful episodes with Form2Fit are also uniformly distributed over the workspace, which further confirms that its failures come largely from mismatching pick-and-place actions, rather than its inability to spatially interpolate or extrapolate.

In the low data regime (with 10 demonstrations), GT-State MLP seems to overtrain to correlations that do not necessarily enable task success, \eg that picking locations are conditioned on the location of the target fixture (see distribution of failed pick locations and fixture locations). By dividing GT-State MLP into two steps -- one for picking and one for placing, we see less issues with incorrect correlations. However, the inability to extrapolate too far beyond the distribution of training data still remains (\eg see failure testing configurations of fixture for GT-State MLP 2-Step with 100 demonstrations, which also caused an incorrect pick point prediction).

\subsection{Ablation Studies}

Tab. \ref{table:ablation} also reports the performance of various ablations of Transporter Networks:

% TODO: redo conditioning placing model on pick results.
\textbf{No-Transport} uses two FCNs: one to predict a picking heatmap, another to predict placing heatmaps (one per discretized rotation). The transport model is not used here, hence there is no pick-conditioned placing. For simpler tasks, such as place-red-in-green, this could suffice if the picking model predicts higher values on red pixels, and the placing model predicts higher values on green pixels. Or, for block-insertion, if the picking model detects the same location on the block, and the placing model detects the same corresponding location and orientation in the fixture. However, it is much more challenging to execute tasks that require additional visual reasoning, such as aligning unseen box corners. For palletizing and packing boxes, the learned policies are generally capable of placing a few of the boxes into the target zones, but fail to align them to each other tightly to achieve higher completion scores above 55\%.
	
\textbf{Per-Pixel-CE-Loss} is a variant of the Transporter Network, where the cross-entropy loss over the entire image (representing a single probability distribution) is replaced with per-pixel cross-entropy, where each pixel represents its own probability distribution. During training, for every demonstration pick-and-place: the picking and placing pixels unraveled from the data are chosen as positive samples for the models respectively, while a fixed number of 100 random negative pixels are sampled from everywhere else. In general, our experiments suggest that the model i) takes more training iterations to converge, and ii) requires more hyper-parameter tuning to balance the ratio of positives to negatives to regularize the loss over time.

\begin{table}[h]
  \setlength\tabcolsep{2.3pt}
  \centering
  \scriptsize
  \begin{tabular}{@{}lcccccccccccccccccccc@{}}
  \toprule
  & \multicolumn{4}{c}{block-insertion} & \multicolumn{4}{c}{place-red-in-green} & \multicolumn{4}{c}{towers-of-hanoi} & \multicolumn{4}{c}{align-box-corner} & \multicolumn{4}{c}{stack-block-pyramid} \\
  \cmidrule(lr){2-5} \cmidrule(lr){6-9} \cmidrule(lr){10-13} \cmidrule(lr){14-17} \cmidrule(lr){18-21}
  Method & 1 & 10 & 100 & 1000 & 1 & 10 & 100 & 1000 & 1 & 10 & 100 & 1000 & 1 & 10 & 100 & 1000 & 1 & 10 & 100 & 1000 \\
  \midrule
  Per-Pixel-CE-Loss                 & 89.0 & 88.0 & 91.0 & 90.0 & 87.5 & 100 & 100 & 100 & 29.6 & 57.0 & 57.9 & 64.6 & 46.0 & 65.0 & 73.0 & 77.0 & 23.0 & 28.0 & 33.7 & 36.0 \\
  No-Transport                      & 5.0 & 8.0 & 6.0 & 8.0 & 81.5 & 95.7 & 100 & 100 & 2.1 & 3.6 & 4.7 & 3.3 & 9.0 & 7.0 & 11.0 & 15.0 & 8.0 & 13.2 & 16.0 & 25.8 \\
  \midrule
  & \multicolumn{4}{c}{palletizing-boxes} & \multicolumn{4}{c}{assembling-kits} & \multicolumn{4}{c}{packing-boxes} & \multicolumn{4}{c}{manipulating-rope} & \multicolumn{4}{c}{sweeping-piles} \\
  \cmidrule(lr){2-5} \cmidrule(lr){6-9} \cmidrule(lr){10-13} \cmidrule(lr){14-17} \cmidrule(lr){18-21}
  & 1 & 10 & 100 & 1000 & 1 & 10 & 100 & 1000 & 1 & 10 & 100 & 1000 & 1 & 10 & 100 & 1000 & 1 & 10 & 100 & 1000 \\
  \midrule
  Per-Pixel-CE-Loss                 & 40.3 & 41.5 & 39.1 & 42.7 & 12.8 & 48.0 & 50.0 & 52.6 & 44.8 & 58.1 & 58.9 & 68.1 & 4.9 & 29.7 & 28.8 & 38.9 & 30.3 & 29.9 & 37.5 & 39.6 \\
  No-Transport                      & 44.3 & 50.4 & 53.6 & 54.1 & 0.6 & 4.0 & 1.8 & 2.8 & 11.3 & 40.1 & 42.3 & 53.5 & 33.5 & 35.3 & 38.8 & 40.0 & 13.7 & 18.8 & 28.8 & 38.1  \\
  \bottomrule
  \end{tabular}
  \vspace{0.5em}
  \caption{\scriptsize\textbf{Ablative comparisons.} Task success rate (mean \%) vs. \# of demonstration episodes (1, 10, 100, or 1000) used in training.}
  \label{table:ablation}
  \vspace{-2em}
\end{table}

\begin{wrapfigure}{r}{0.35\textwidth}
  \vspace{-2.4em}
  \begin{center}
    \includegraphics[width=0.35\textwidth]{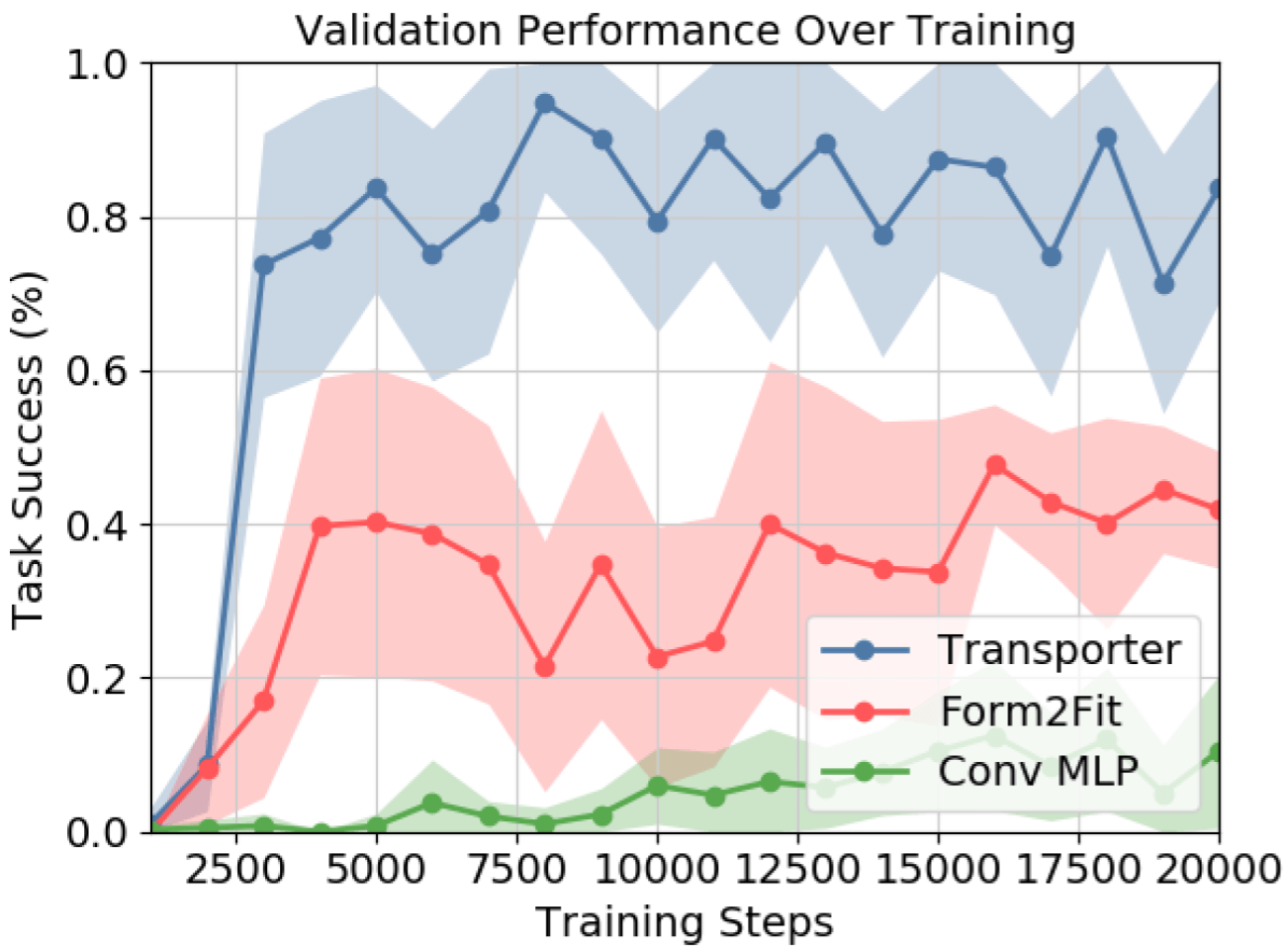}
  \end{center}
  %\caption{\textbf{Translation-only.} Caption here.}
  %\label{fig:training-plot}
  % https://docs.google.com/drawings/d/1JYVa5ULPgR1m1fp6HE7YpdocD5cStXfcDwk8Jp6VdlE/edit
  \vspace{-4em}
\end{wrapfigure}

\subsection{Training Convergence}

We train Transporter Networks with Adam \cite{kingma2014adam}, using fixed learning rates of $10^{-5}$.
Our models are trained from scratch (\ie random initialization) in Tensorflow.
On the right, we show an example convergence plot over training from 100 demonstrations of manipulating rope.
Compared with image-based baselines, Transporter Networks generally converge faster -- within a few thousand training iterations, \ie one or two hours of wall-clock time.

\subsection{Visualizing Transporter Network Predictions}
\label{sec:modes}

\begin{figure}[t]
\centering
  \includegraphics[width=\textwidth]{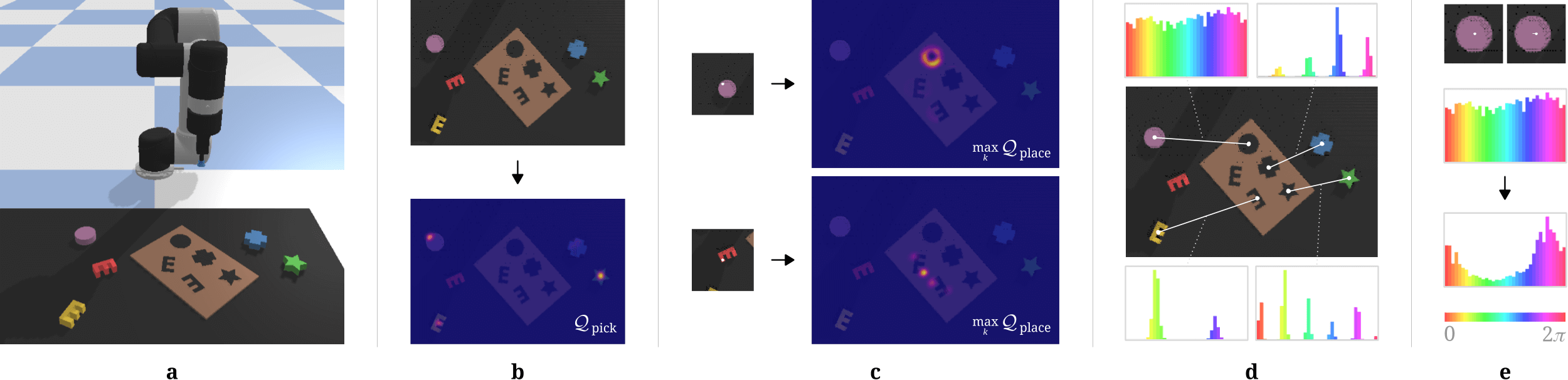}
  \caption{\textbf{Visualizing Transporter Network Predictions.} For the kit assembly task (a), Transporter Network picking predictions show 3 dominant modes on 3 different objects (b) which are not necessarily centered on the object (which may reflect biases learned from the distribution of training data). If a pick is off-center \eg on the circle block, (c, top), the Transporter Network correctly makes placing predictions along the edges of the target placement location -- which is visualized as a ring on $\mathcal{Q}_\textrm{place}$ maxed over rotations. If a pick is on the corner of the unseen E-shaped object (c, bottom), then two dominant modes lie on the correct corresponding locations of their target placements -- but also with false positive modes from geometric symmetries in the letter E. We also show modes of predicted placing values over rotations for pick-and-place locations centered on the object and fixture, which appear uniform for circle, 4 modes for the cross shape, 5 modes for star shape, 2 modes for the unseen E-shaped block (dominant one being correct). We also visualize how the predicted placing values over rotations change from uniform, to unimodal, when shifting one pixel to the right from the center of the circle.}
  \label{fig:modes}
  % https://docs.google.com/drawings/d/1KdSpIyPJdLSLeSK7ayH65ZSyGxzswsHX2edaZ_aqywg/edit
\end{figure}

Fig. \ref{fig:modes} visualizes Transporter Network dense pixel-wise predictions of $\mathcal{Q}_\textrm{pick}$ and $\mathcal{Q}_\textrm{place}$ for the kit assembly task (trained with 1,000 demonstrations), and how the predicted modes change with respect to the locations of objects, their shapes, and conditioned picks. These visualizations show that due to rotation and translation equivariance, our method can quickly learn modes that reflect symmetries in the data.

These visualizations also show an interesting failure mode, in which predicted placements of the unseen E-shaped object returns several 180 degree flipped false positives (see Fig. \ref{fig:modes}c). This is because the learned features from this task rely strongly on local geometry. Since the E shape is a new unseen object, a flipped E looks similar enough to return a false positive signal there (though not enough to be the top mode).

\subsection{Real-World Experiments Details}
\label{sec:real-experiments-details}

\begin{wrapfigure}{r}{0.6\textwidth}
  \vspace{-1.8em}
  \begin{center}
    \includegraphics[width=0.6\textwidth]{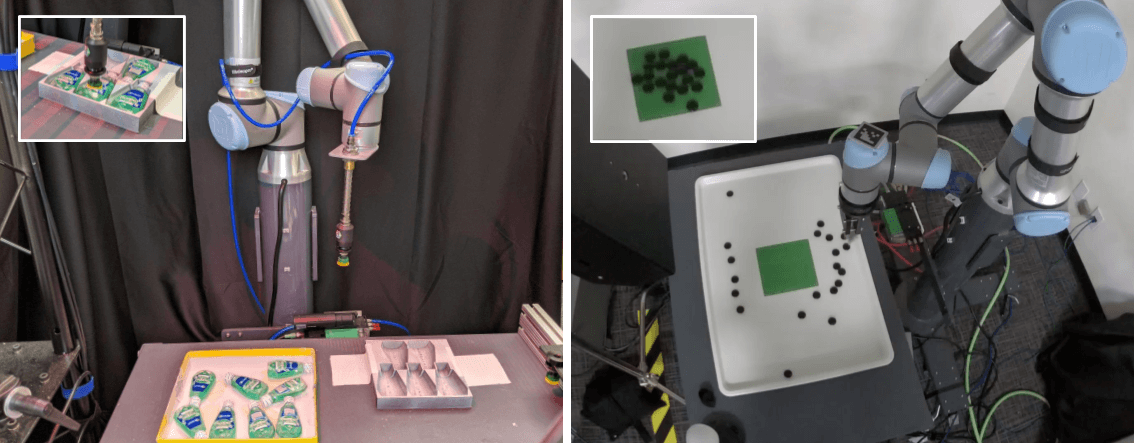}
  \end{center}
  \vspace{-0.5em}
  \caption{\textbf{Real-world tasks.} \emph{Left:} assembling kits of mouthwash bottles, \emph{right:} pushing piles of small Go pieces.}
  \label{fig:real-tasks}
  %https://docs.google.com/drawings/d/18ggORz2NIJnCssUzA4phWTvZ67bpUP9RAwpww7AYvU0/edit
  \vspace{-1.6em}
\end{wrapfigure}

To demonstrate the real-world performance of Transporter Networks, we apply our approach to two challenging real-world tasks: assembling kits of mouthwash bottles with pick and place, and sweeping piles of small Go pieces with pushing. These tasks are shown in Fig. \ref{fig:real-tasks}. 

% 98.3 weeping, 300 s

% 

For kit assembly, we sequentially pick 5 small mouthwash bottles from a ``bin" of identically shaped items and place each into a ``kit" with 5 corresponding indentations for placement. This is a challenging task due to the multi-modality and ambiguity of the pick location (many identical items in the bin to pick from), as well as the tight millimeter tolerances of the destination kit. It is designed to mimic a production robotic task of placing items into a blister pack for consumer packaging. Kit assembly and disassembly are treated as separate tasks -- both use the same Transporter Network architecture with autonomous switching between the two tasks at inference time. As shown in Table \ref{table:real}, our model achieves 98.9\% on kit assembly, trained with 8,141 pick and place actions recorded from human demonstrations. Picking success and disassembly success are both above 99\%.

For sweeping, we sequentially push piles of small Go pieces into a target goal zone, marked with green tape. Upon completion of the task, the robot autonomously resets with a scripted scrambling motion. This task is challenging as it tests the capacity of the model to sequence actions using closed-loop visual feedback, and to do it efficiently. As shown in Table \ref{table:real}, our model achieves 98.3\% on this task, trained with 6,759 pushing actions recorded from human demonstrations. The average number of sweeps needed to complete the task is 20 $\pm$ 2.5 sweeps, with the fastest run being 16 sweeps and the slowest 25. For reference, the average for human teleoperation is 18.1 sweeps.

% Task B involves picking from 9 uniquely shaped and colored wooden toys and placing them into a kit with corresponding unique indentation. For this task, the policy must learn the object to destination indent correspondence (which shape goes in which indent) and infer a precise placement pose in order to fit the shape.

To train our system for both tasks, we use human demonstration data captured by a remote teleoperation interface. The sub-optimal, biased and noisy distribution of human demonstrations presents a further challenge for robotic policy learning.
%\cite{}. \jonathan{TODO: need a cite here?}.  Can't think of one. Probably OK as is.
In addition, 13 teleoperators of varying skill and performance level provide training data, which further increases the variance of ground-truth actions that we train from and subsequently increases the difficulty of the learning task. 

\textbf{Hardware Setup}

% Source for workspace_diagram.pdf
% https://docs.google.com/drawings/d/1x4ugvn3fiJOMD3Mh-PdqRCzUgPCE_LAaLzbcNgGMaWo/edit?usp=sharing
\begin{wrapfigure}{r}{0.43\textwidth}
  \vspace{-1.8em}
  \begin{center}
    \includegraphics[width=0.43\textwidth]{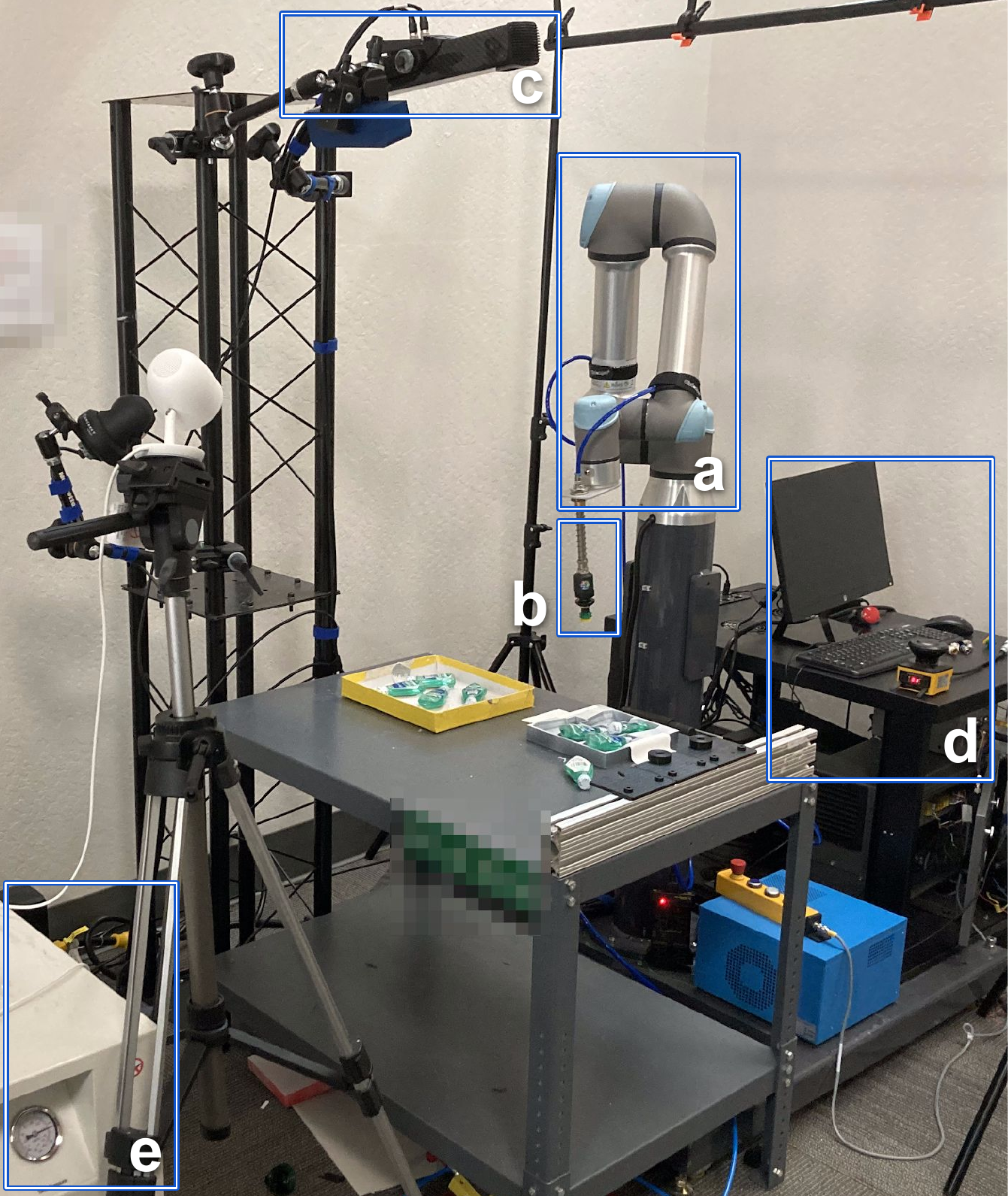}
  \end{center}
  \vspace{-0.5em}
  \caption{\textbf{Hardware setup.} a) UR5e Robot, b) Piab suction cup, c) Photoneo camera, d) Linux PC, and e) air compressor and vacuum generator.}
  \label{fig:hardware-setup}
  \vspace{-2em}
\end{wrapfigure}
The hardware setup for kit assembly is shown in Fig~\ref{fig:hardware-setup}. Our experiments make use of 2 Universal Robots UR5 workstations (one for kit assembly and one for sweeping), each with an industrial Linux PC, vacuum system, Piab suction cup or brush end-effector, and depth camera (statically mounted above the workstation).
% Our system can work with different robots like UR5, UR10 several different Fanuc industrial robots of varying sizes as well as Kuka or Franka robots.
% jonathan: Cannot make the above statement without experiments IMO.
To test the robustness of our method to diverse data sources, we use high quality depth camera observations from a Photoneo PhoXi Model M for kit assembly, and those from low cost consumer-grade Azure Kinect for sweeping. The Photoneo camera supplies depth (with 0.1mm rated depth precision) and a greyscale IR image, both at $1032\times772$ resolution. The Kinect camera supplies depth as well as a color RGB image at $1280\times720$ resolution.

To calibrate the camera within the robot coordinate frame we use a two step procedure. Firstly, the intrinsics of the camera are calibrated by capturing multiple images of a large flat QR-Code panel in varying orientations, and we use OpenCV to calculate the camera intrinsics.
% We calibrate the (area-scan-camera(this is the industrial term) || 2d camera || matrix camera || RGB/Mono camera) according to the OpenCV camera model and also determine polynomial  depth distortion offset by assuming flatness of the calibration board.
% Ideally these intrinsics will stay the same and do no not often have to be repeated in production setup.
Secondly, to calibrate extrinsics we attach QR code tags to the UR5 wrist joint and we capture multiple images of the robot in random end-effector poses. These images are then used to solve for the position of the robot base as well as the offset of the QR Tag to its respective joint with a stochastic optimization procedure.
% The cameras can either be configured as a workcell-mounted setup separate from the robot or connected to any of the robot joints which enables very flexible workcell configurations.
% Our systems are completely capable to have the camera mounted at arbitrary angles and pick objects from a bin at any orientation.
% To account for any changes that happen to the work-cell we can run this calibration procedure automatically to avoid slowly degrading calibrations of time, when for example camera mounts are not perfectly stable and drift slowly. 
% jonathan: Is it worth mentioning the donut calibration? Probably not.

A Linux system embedded with the robot gathers robot and camera data and forwards it to the teleoperator over the internet. Any commands passed to the robots are queued up to be processed one-by one. All data is logged in real-time via network to a cloud storage database for model training and evaluation.

% In industrial applications most systems are picking from a flat surface and placing to a flat surface such as from one conveyor to another, these types of configurations are very common in industrial automation for a long time and have been solved successfully with traditional 2D machine vision approaches like pattern matching and offset correction. When 3D cameras were needed in the past, the solutions were mostly custom and have been very difficult to fully automate and maintain. With transporter networks we present a solution framework that performs well in the domain where classical 2D vision applications are used today, but we have full capability to seamlessly solve more difficult 3D picking problems like bin to picking and accurate placement as intermediate 2.5 d applications where the 2D image is mostly used for location and orientation, but a depth measurement is used for height.

% Should we add schematic drawings here ?
% Jonathan: I don't think so. I think the figures in the main body of the text are descriptive enough. One option would be to:

\textbf{Data Collection}

\begin{figure}[t]
\centering
  \includegraphics[width=\textwidth]{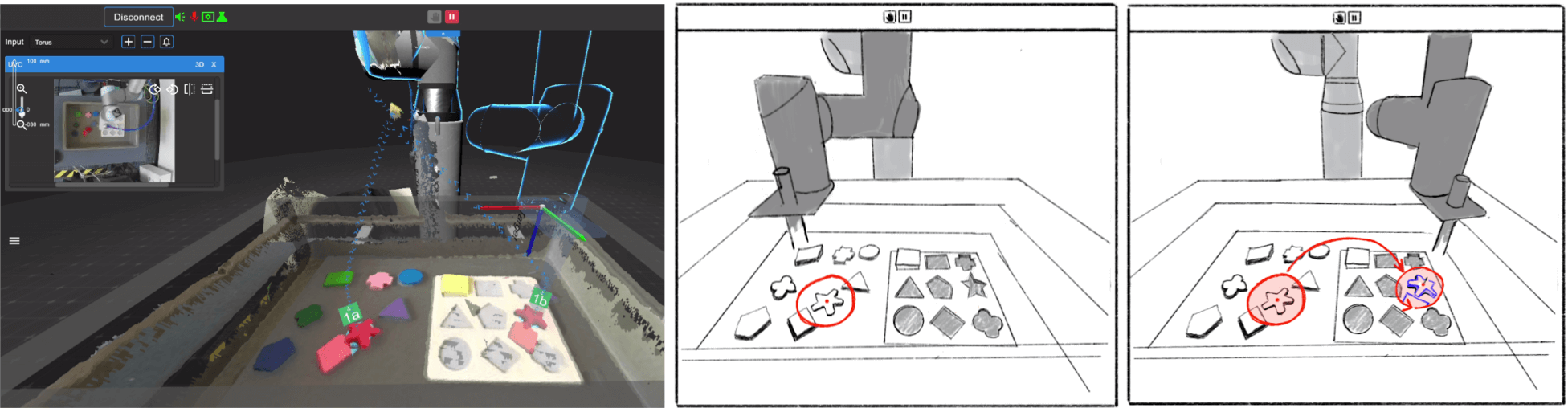}
  \caption{\textbf{Data Collection UI.} \emph{Left:} Unity-based UI. \emph{Middle:} The operator selects the area where the object is to be picked up. \emph{Right:} The operator can adjust the placement of the object and rotate it. The depth image around the pick-up point is overlaid and helps the operator align the placement. % "transported" over that is why the star is blue "it visualizes the fit" the operator can rotate the fit.
  }
  \label{fig:data-collection}
  % \vspace{-1em}
  % https://docs.google.com/drawings/d/17h-sxKHBPpYazFeN5-YexSpMpNedHxHxxiQhPqkUH9s/edit
\end{figure}
%\andy{TODO: cleanup description} \jonathan{I had a go at it. However, this content is already in the text below (in slightly more verbose form). Not sure if we need to repeat it in the caption.} andy: looks great!

% tompson_20201026: This is rendering a littler weirdly in the pdf. Not sure why latex is breaking this paragraph when the figure is on the previous page? :-(
We designed a 2D User-Interface (UI) scripted in Unity~\cite{unity2018unity} in order to control the robot workstations remotely. An overview of this UI is shown in Fig~\ref{fig:data-collection}. Features of this UI include: configuration of the robot workcell, movement and actuation of the suction end-effector, configuration of parameterized action trajectories, calibration of the robot and control of the robot using the same action parameterization used to train our model.

A standard teleoperation session is as follows. The operator views a 3D rendering of the workstation depth data. The synthetic camera view can be rotated and translated using mouse movements as desired by the operator. The operator determines with mouse clicks which object to pick by selecting the intended pick-up point on the rendered depth geometry. A region of the point cloud around the pick location is then ``attached" to the mouse cursor, where it is rendered transparent and overlaid on the existing depth (illustrated in Fig~\ref{fig:data-collection}). The operator is able to translate this geometry with movements of the mouse and the single rotational degree of freedom is controlled by the mouse wheel. Once the operator moves the object into the desired location, an additional mouse click will select the 3DOF place location. This interface allows the operator to visualize the intended 3D configuration of the pick-and-place operation before the action is executed. The discrete teleoperator pick-and-place locations are then translated into dense robot trajectories using a Cartesian controller, checked for collisions and other safety constraints, and subsequently executed on-robot. The user then repeats this process until the task is complete.

%\jonathan{TODO: Add UI screenshots?}
%\andy{added (done by Stefan)}
\begin{figure}[t]
\centering
  \includegraphics[width=\textwidth]{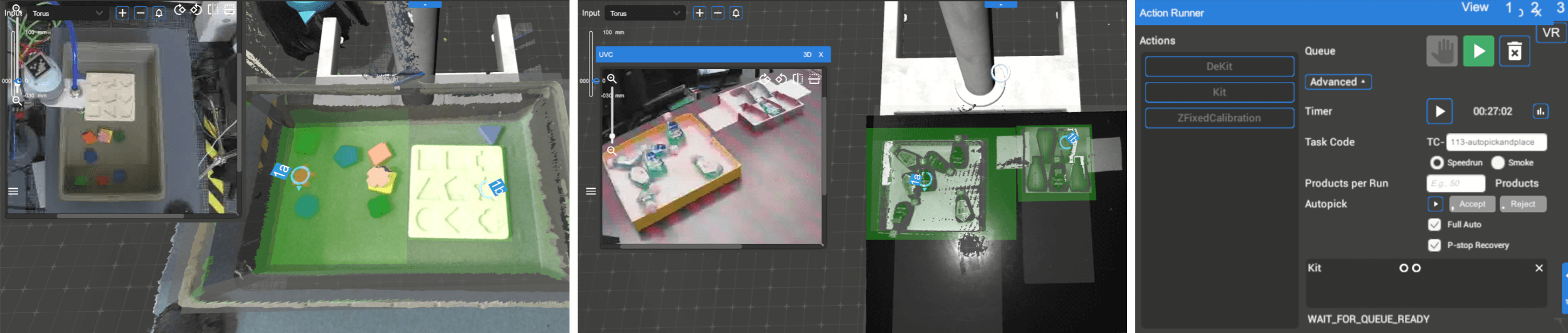}
  \caption{\textbf{Autonomous Mode UI.} User inputs the task to execute (by unique task code), and presses the play button to begin autonomous execution.}
  \label{fig:data-collection-autopick}
  % https://docs.google.com/drawings/d/17h-sxKHBPpYazFeN5-YexSpMpNedHxHxxiQhPqkUH9s/edit
\end{figure}

Note that we use the same data pipeline and motion primitives when executing our Transporter Networks model as is used during human teleoperation; the human UI interaction is replaced with our inference results, and additional logic is added to switch between assembly and disassembly tasks. Autonomous mode is executing within the standard UI by first inputting the task to execute (by unique task code identifier), and pressing start. The UI for this (and visualization of inference results) is shown in Fig~\ref{fig:data-collection-autopick}.

Using the above system we collected approximately 8,141 pick-and-place actions from 13 human operators in order to train the kit assembly model, an additional 2,300 actions for validation of model performance and hyper parameter tuning, and 6,759 pushing actions to train the sweeping model.

\textbf{Success Detection}
\label{sec:success-detection}

% Jonathan: We can't add this content until we use the trained success detector somewhere in the experiments, otherwise reviewers will ask for percent-success numbers. Lets continue to work on it and then add it in the v2 draft / camera ready if it's used.

% maria here
For the on-robot evaluation in Section~\ref{sec:real-world-experiments}, we manually label robot actions for success or failure. However this process is time consuming and expensive. Additionally, our Transformer Network does not estimate when a multi-action kitting episode terminates (either successfully or unsuccessfully). So in order to perform fully-autonomous pick-and-place without manual success labelling, an automatic task-level success signal was implemented. We trained a ResNet50 classification model - pretrained on imagenet classification - to perform 3-way classification in order to detect full, empty or partial kits. We trained the classifier on 4848 training samples and evaluated on 128 validation set samples from a unique camera angle. Mean classification accuracy on the validation set was \textbf{96.9\%}. Upon completion of a pick-and-place attempt, classifier inference on the RGB image produces a success metric, which marks the successful completion of a Kitting (or emptying the kit) episode. This allows the fully autonomous mode of the system to alternate between tasks (kitting and emptying) upon completion.

\end{document}